\documentclass[sigconf,noacm]{acmart}

\usepackage[]{algorithm,array}
\usepackage{algpseudocode}
\usepackage{pifont}
\usepackage{pgfplots}
\usepackage{listings}
\usepackage{subcaption}
\usepackage{float}

\usepackage{tikz}
\tikzset{box/.style={draw, thick, minimum width=1cm, minimum height=0.7cm}}
\usepackage{circuitikz}
\usetikzlibrary{positioning}
\usepackage{multirow}
\newtheorem{problem}{Problem}

\usepackage[
    skins,
    breakable
]{tcolorbox}

\usepackage{lipsum}
\usepackage{url}

\AtBeginDocument{%
  }

\begin{document}

\title{Continuous Fairness On Data Streams}

\author{Subhodeep Ghosh}
\email{sg2646@njit.edu}
\affiliation{
  \institution{NJIT}
  \country{USA}
  }

\author{Zhihui Du}
\email{zhihui.du@njit.edu}
\affiliation{%
  \institution{NJIT}
  \country{USA}
  }

   \author{Angela Bonifati}
\email{angela.bonifati@univ-lyon1.fr }
\affiliation{%
  \institution{Lyon 1 University}
  \country{France}
  }

  \author{Manish Kumar}
\email{manishk@iitrpr.ac.in}
\affiliation{%
  \institution{IIT Ropar}
  \country{India}
  }

   \author{David A. Bader}
\email{bader@njit.edu}
\affiliation{%
  \institution{NJIT}
  \country{USA}
  }

\author{Senjuti Basu Roy}
\email{senjutib@njit.edu}
\affiliation{%
  \institution{NJIT}
  \country{USA}
  }


\renewcommand{\shortauthors}{Ghosh et al.}


\begin{abstract}

We study the problem of enforcing continuous group fairness over windows in data streams. We propose a novel fairness model that ensures group fairness at a finer granularity level (referred to as block)  within each sliding window. This formulation is particularly useful when the window size is large, making it desirable to enforce fairness at a finer granularity. Within this framework, we address two key challenges:  efficiently monitoring whether each sliding window satisfies block-level group fairness, and 
reordering the current window as effectively as possible when fairness is violated. To enable real-time monitoring, we design sketch-based data structures that maintain attribute distributions with minimal overhead. We also develop optimal, efficient algorithms for the reordering task, supported by rigorous theoretical guarantees. Our evaluation on four real-world streaming scenarios demonstrates the practical effectiveness of our approach.  We achieve millisecond-level processing and a throughput of approximately $30{,}000$ queries per second on average, depending on system parameters. The stream reordering algorithm improves block-level group fairness by up to $95\%$ in certain cases, and by $50$--$60\%$ on average across datasets. A qualitative study further highlights the advantages of block-level fairness compared to window-level fairness.

\end{abstract}

\maketitle

\section{Introduction}
Many real-world applications inherently generate data streams rather than static datasets - for example, financial market tickers, performance metrics in network monitoring and traffic management, server logs, content recommendation engines, and user click streams in web analytics and personalization systems~\cite{babu2001continuous,golab2003processing}. Due to the continuous and evolving nature of data streams, querying them requires continuous queries, i.e. queries that are executed persistently over time and return updated results as new data arrive.
A fundamental concept in stream processing are the sliding windows, allowing to analyze continuously flowing data in real time.

    \begin{figure}[htbp]
    \centering
    \includegraphics[width=\linewidth,height=6cm]{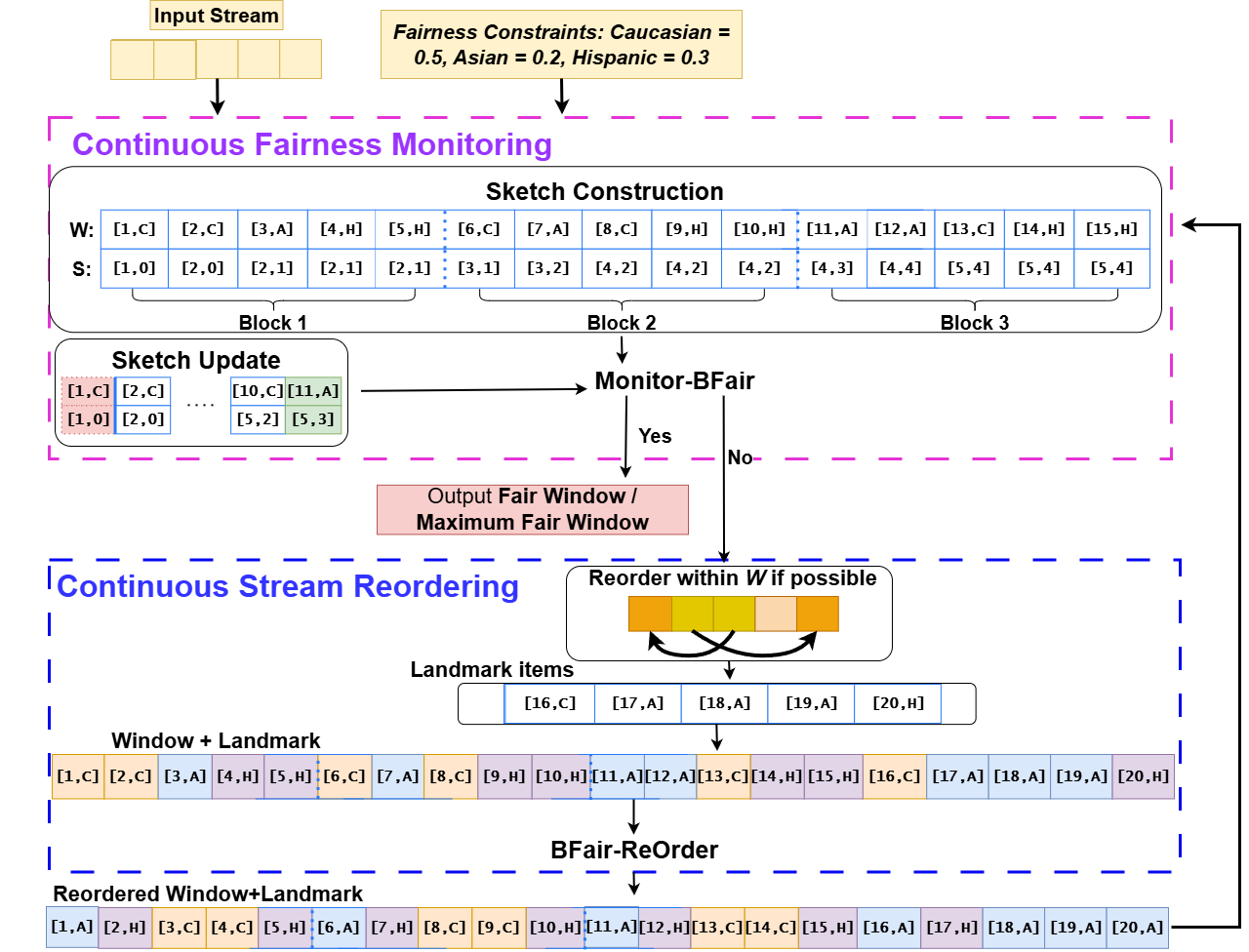}
    \caption{\small The proposed framework consists of two tightly coupled components. For the first ever window, it constructs a compact \emph{fairness sketch} that summarizes block-level counts; for each subsequent window, this sketch is incrementally updated as the window slides. This sketch is used to perform \emph{continuous fairness monitoring}, enabling efficient verification of block-level group fairness constraints within each window. When a window violates fairness, the framework first attempts to perform \emph{continuous stream reordering} using only the items in the current window. If no feasible reordering exists, it reads an additional set of landmark items and jointly reorders the window and landmarks. The reordering objective is to maximize the number of unique fair blocks across the current window and the additional windows induced by the landmark items.}
    \label{fig:framework}
\vspace{-0.1in}
\end{figure}

In parallel, fairness~\cite{binns2020apparent} in algorithmic decision-making has emerged as a critical issue in data management, especially in high-stakes domains such as hiring, lending, and law enforcement~\cite{pitoura2022fairness,stoyanovich2020responsible,garcia2021maxmin,kuhlman2020rank,wei2022rank}. A substantial body of research has focused on {\em group fairness}, which seeks to ensure that algorithmic outcomes do not systematically disadvantage particular demographic groups. However, most of this work assumes access to static datasets, where fairness constraints are applied once on a fixed input. Despite the widespread relevance of these streaming scenarios in real-world applications, the study of fairness in streaming settings remains significantly underexplored.

 \noindent {\bf Motivating Application - Online Content Recommendation}\label{ex}

In a news or video streaming platform that continuously recommends content in real time, items arrive from creators across different ethnic groups (e.g., Caucasian, Asian, Hispanic). The system displays this content in paginated form, where each sliding window of recommendations is divided across multiple pages.

Even when the overall window includes a balanced mix of content from all groups, the page-wise distribution may still be skewed — for example, the first few pages may predominantly feature Caucasian creators, while later pages contain more balanced or diverse representation.

Since users’ attention typically drops sharply across pages — with most engagement concentrated on the first one or two — this ordering bias results in unequal visibility and exposure. Thus, ensuring proportional representation across an entire window is insufficient; fairness must also be enforced within and across pages, accounting for both the presentation order and attention decay patterns.

\noindent {\bf Contributions.} 
In this paper, for the first time we study group fairness on streaming data. 
We introduce a novel fairness model that partitions each sliding window into multiple blocks and enforces group fairness to satisfy proportion of different protected attribute values at the block level (Figure~\ref{fig:framework} contains three such blocks). 
These blocks, naturally defined by the application (e.g., page), provide a localized and flexible alternative to traditional notions of fairness that operate only at the window level or based on prefixes. When an application restricts each window to a single block, our formulation naturally reduces to the classical window-level group fairness model.


To support this framework, we study two interrelated problems: {\bf Continuous Fairness Monitoring} and the {\bf Continuous Stream Reordering Problem}. The monitoring module continuously checks whether fairness constraints hold within each block of every sliding window (see Continuous Fairness Monitoring in Figure~\ref{fig:framework}). When a violation is detected, the system first attempts to reorder the items within the current window to restore fairness. If reordering alone cannot resolve the issue, the decision making pauses - it reads a small number of additional items—referred to as landmark bits—which effectively extend the current stream and create new sliding windows. Conceptually, these landmarks represent a limited “buffer time” that the application is willing to tolerate before producing output (see Continuous Stream Reordering in Figure~\ref{fig:framework}). The framework then performs reordering over both the current window and the landmark bits. We formalize this task as the {\em Continuous Stream Reordering Problem}, aimed at maximizing the number of distinct fair blocks across both current and newly spawned slid windows. Once the reordering is done, {\bf Continuous Fairness Monitoring} outputs the fair window.

We define a {\bf sketch-based} data structure to pre-process every window and an Algorithm {\tt Monitor-BFair} for answering {\em Monitoring Continuous Fairness}. The sketch compactly encodes protected attribute distributions within each block of a sliding window, enabling exact and efficient fairness evaluation. The sketch requires only linear space, supports fast incremental updates, and avoids full recomputation as the window slides. We provide analytical guarantees on correctness, running time, update efficiency, and space complexity, demonstrating its scalability and low-latency performance for high-throughput streams. 

We propose an {\bf efficient and optimal} algorithm {\tt BFair-ReOrder} to the {\bf Continuous Stream Reordering} problem, which incorporates both the current window and the landmark items to maximize the number of \emph{unique fair blocks} (where two blocks are considered unique if they have at least one uncommon item). Each fair block must satisfy specified group-wise representation constraints, and the goal is to maximize the number of such blocks not only within the current window but also across future windows defined by landmark positions. 
The key idea of the algorithm is to construct a specific permutation of the content, called an \emph{isomorphic stream} (see details in Sec.~\ref{sec:reorder}), which satisfies the fairness requirement consistently within its blocks. We present theoretical analyses showing optimality and computational overhead.

We conduct extensive experiments using four large-scale real-world datasets by simulating streaming environment. The results align with our theoretical analysis, confirming the optimality of our solutions. The proposed framework processes queries in fractions of milliseconds, achieving an average throughput of 30,000 queries per second (depending on parameters), with minimal memory usage and high update efficiency. As no existing work addresses our problem setting, we design appropriate baselines. Our framework outperforms them by several orders of magnitude.
To summarize, the paper makes the following contributions:
\vspace{-0.05in}
 \begin{itemize}
\item {\bf Block-level Fairness Model.} We propose a fairness model that partitions each sliding window into multiple blocks, enforcing localized group-fair exposure. This generalizes window-level fairness and offers a flexible alternative to prefix-based notions.
    
\item {\bf Continuous Fairness Monitoring and Stream Reordering.} We formalize the problem of  {\em Monitoring Continuous Fairness} via a sketch-based algorithm {\tt Monitor-BFair} along with the {\em Continuous Stream Reordering Problem}, solved by our optimal algorithm {\tt BFair-ReOrder} that constructs an {\em isomorphic stream} to maximize unique fair blocks.
  
\item {\bf Theoretical Guarantees.} We provide formal analysis of correctness, optimality, space, and time complexity. {\tt B-FairQP} ensures linear space and fast incremental updates, while {\tt Block-FairReOrder} is proven optimal and efficient.  

\item {\bf Experimental Validation.} On four large-scale real-world datasets, our framework achieves sub-millisecond running times, throughput up to 30,000 queries/second, and minimal memory usage, outperforming baselines by several orders of magnitude.
\end{itemize}

\section{Data Model and Framework}
We first present our data model, followed by our proposed framework, based on which we formalize the studied problems. We start by introducing an example  based on the motivating application discussed in the introduction.

\begin{example}\label{ex1}

\noindent A sliding window contains $15$ items divided in 3 blocks (pages)- each item contains content generated by individuals with a protected attribute ethnicity. Without loss of generality, imagine, there are 3 different ethnic groups, shown in Figure~\ref{fig:windowblock}. There are two representative fairness constraints - shown in Table~\ref{queryconstraints}.
\begin{figure}[!htbp]
\centering
\begin{tikzpicture}[scale=0.8]
    \scriptsize
    \node at (0, 0.5) {$B_1$:};
    \foreach \i in {1,2,3,4,5} {
        \draw (\i, 0) rectangle (\i + 1, 1);
    }
    \node at (1.5, 0.5) {$[1,\emph{C}]$};
    \node at (2.5, 0.5) {$[2,\emph{C}]$};
    \node at (3.5, 0.5) {$[3,\emph{A}]$};
    \node at (4.5, 0.5) {$[4,\emph{H}]$};
    \node at (5.5, 0.5) {$[5,\emph{H}]$};
\end{tikzpicture}

\vspace{6mm}

\begin{tikzpicture}[scale=0.8]
    \scriptsize
    \node at (0, 0.5) {$B_2$:};
    \foreach \i in {1,2,3,4,5} {
        \draw (\i, 0) rectangle (\i + 1, 1);
    }
    \node at (1.5, 0.5) {$[6,\emph{C}]$};
    \node at (2.5, 0.5) {$[7,\emph{A}]$};
    \node at (3.5, 0.5) {$[8,\emph{C}]$};
    \node at (4.5, 0.5) {$[9,\emph{H}]$};
    \node at (5.5, 0.5) {$[10,\emph{H}]$};
\end{tikzpicture}

\vspace{6mm}

\begin{tikzpicture}[scale=0.8]
    \scriptsize
    \node at (0, 0.5) {$B_3$:};
    \foreach \i in {1,2,3,4,5} {
        \draw (\i, 0) rectangle (\i + 1, 1);
    }
    \node at (1.5, 0.5) {$[11,\emph{A}]$};
    \node at (2.5, 0.5) {$[12,\emph{A}]$};
    \node at (3.5, 0.5) {$[13,\emph{C}]$};
    \node at (4.5, 0.5) {$[14,\emph{H}]$};
    \node at (5.5, 0.5) {$[15,\emph{H}]$};
\end{tikzpicture}
\caption{\small A sliding window of $15$ items (contents) with $3$ blocks, consisting of contents from three ethnic groups $[\emph{C}$ (Caucasian), $\emph{A}$ (Asian), $\emph{H}$ (Hispanic)].}
\label{fig:windowblock}
\end{figure}
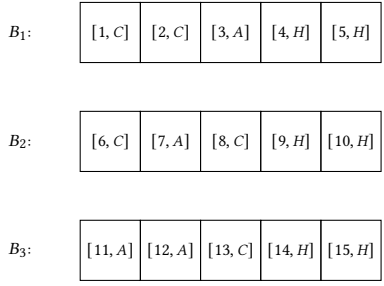

\begin{table}[!htbp]
\centering
\scriptsize
\begin{tabular}{|p{0.9cm}|p{0.9cm}|p{0.9cm}|p{0.9cm}|}
\hline
\textbf{Attr.} & \textbf{Value} & \textbf{C1} & \textbf{C2} \\ \hline
\multirow{3}{*}{Ethnicity} 
& {\em C}  & 0.3 & 0.5 \\ \cline{2-4}
& {\em A}  & 0.3 & 0.2 \\ \cline{2-4}
& {\em H}  & 0.4 & 0.3 \\ \hline
\end{tabular}
\caption{\small Two fairness constraints.}
\vspace{-0.1in}
\label{queryconstraints}
\end{table}
\end{example}

\vspace{-0.1in}
\subsection{Data Model}\label{sec:dm}
\smallskip \noindent {\bf Attribute denoting fairness.}
A protected attribute $\mathcal{A}$ has $\ell$ possible values. Using Figure~\ref{ex}, {\em Ethnicity} is one of these attributes  with  three possible values, {\em C, A, H}.  \\
\vspace{-0.05in}
\smallskip \noindent {\bf Count based sliding window.}
A count-based sliding window in stream processing performs computations over the most recent fixed number of data items and advances by a specified number of items, referred to as the {\em slide}. We use $W$ to denote a sliding window containing $|W|$ items. Essentially, the window maintains the latest $|W|$ elements of the data stream and updates to a new window each time a new item arrives. As data continuously flows in, we denote the $y$-th such window as $W_y$. Unless otherwise specified, the default slide size is $1$.

Let $$W_y = \{\, t_{y},\; t_{y+1},\; \ldots,\; t_{y + |W| - 1} \,\}$$
denote the $y$-th sliding window, which contains $|W|$ consecutive items from the data stream. 

For example, 
$W_1 = \{ t_1, \ldots, t_{|W|} \}$, 
$W_2 = \{ t_{2}, \ldots, t_{|W|+1} \}$, and so on. 

Using Example~\ref{ex1}, each sliding window $W$ contains $15$ items. Window $W_1$ contains items $t_1,...t_{15}$, whereas, Window $W_2$ contains items $t_2,...t_{16}$, and so on.

\smallskip \noindent {\bf Data item.}
A data item $t$ in a sliding window $W$ is a pair $t=<i,\mathit{p}>$, where $i$ is its order in the stream (key), and $\mathit{p}$ represents any specific value of the protected attribute $\mathcal{A}$ that is, $\mathit{p} \in \ell$. 

Using Example~\ref{ex1}, the third video of the stream is created by a contributor of Asian ({\em A}) ethnicity.

\smallskip \noindent {\bf Blocks in a window.}
Each window $W$ is further partitioned into $k$ equal-sized sub-intervals, called \emph{blocks}, 
where the block size is defined as
\[
s = \frac{|W|}{k}.
\]
\noindent

Each block $B_q$ in window $W_y$ contains exactly $s$ consecutive items from that window:
\[
\begin{aligned}
B_q 
&= [\, 
    t_{\,y + (q-1)s},\,
    t_{\,y + (q-1)s + 1},\,
    \ldots,\,
    t_{\,y + q s - 1} 
    \,], \\[3pt]
&\quad q = 1, 2, \ldots, k.
\end{aligned}
\]

Every item  in a window therefore belongs to exactly one block within that window.

Using Example~\ref{ex1}, each sliding window contains $k=3$ blocks, each with $15/3=5$ items. Blocks $B_1$, $B_2$, $B_3$ start at item $1$, $6$, and $11$, respectively for $W_1$. Blocks $B_1$, $B_2$, $B_3$ start at item $2$, $7$, and $12$, respectively for $W_2$. Figure~\ref{fig:windowblock} shows $\{B_1,B_2,B_3\}$ of the first window.

For any item $t_j \in W$, its block index is given by
\[
\text{block}(t_j) = \left\lceil \frac{j}{s} \right\rceil.
\]

The choice of block size $s$ (or equivalently, the number of blocks $k$ per window) is guided by the semantics of the target application.
In streaming or recommendation settings, each block typically corresponds to the smallest granularity—such as a page of displayed items—over which fairness or representational balance is desired. The block granularity provides a natural control knob between fairness sensitivity and computational efficiency.


\smallskip \noindent {\bf Prefix, Suffix of a Block.} 
For  a block $B_q$ in a window $W$ starting at index \( i \) 
\[  
B_q = [t_i, t_{i+1}, \ldots, t_{i+s-1}], \quad \text{for } y \le i \le y + |W| - s.
\] 
its prefix of length \( j \) is:
\[
\text{Prefix}_j(B_q) = [t_i, t_{i+1}, \ldots, t_{i+j-1}], \quad \text{for } 1 \le j \le s.
\]
Its suffix of length \( j \) is:
\[
\text{Suffix}_j(B_q) = [t_{i+s-j}, t_{i+s-j+1}, \ldots, t_{i+s-1}], \quad \text{for } 1 \le j \le s.
\]
As an example, the first block $B_1$'s prefix of length $3$ is items $[t_1,t_2,t_3]$ and the suffix of length $2$ is items $[t_4,t_5]$.

\smallskip \noindent {\bf Landmark Items.} 
Given the current window \( W \), the landmark items consist of \( |\mathcal{X}| \) additional elements stored immediately after the window, starting from position \( |W| + 1 \). Figure~\ref{fig:landmark} shows $5$ such additional items.

The landmark size is application-defined and serves as another design knob.  In many streaming scenarios, a short wait for a few additional items beyond the current window is acceptable, as systems often employ slight buffering to refine fairness or diversity without perceptibly affecting latency.  When no delay is permissible, reordering proceeds immediately using the current window, yielding the best fairness achievable with available data.


\smallskip \noindent {\bf Unique Blocks.} \noindent
A block consists of $s$ consecutive items, starting at item $t$ and ending at item $t+s-1$.  Without loss of generality, two blocks are considered \emph{unique} if each contains at least one item not present in the other. For example, the first block in the current window and the first block in the subsequent (slid) window are unique, as one contains items $[t_1, \ldots, t_5]$ while the other contains $[t_2, \ldots, t_6]$ after the slide.


\smallskip \noindent {\bf Disjoint Blocks.} Blocks that do not overlap with one another are called \emph{disjoint blocks}. If each of these blocks individually satisfies the given fairness criterion, they are referred to as \emph{disjoint fair blocks}.  As an example block $B_1$, $B_2$, $B_3$ are fully disjoint.

\smallskip \noindent {\bf Group Fairness Constraints.}
For the attribute $\mathcal{A}$ with $\ell$ different values, for each $\mathit{p} \in \cup_{i=1}^{l}\mathcal{A}_i $, $f(\mathit{p})$ denotes its required proportion constraint, such that $\Sigma_{\mathit{p}=1}^{\ell} f(\mathit{p})=1$. 

Table~\ref{queryconstraints} shows two different group fairness constraints for the {\em Ethnicity} attribute.

\smallskip \noindent {\bf Fair Sliding Window.}
\noindent
Given a sliding window \( W \) and a desired proportion function \( f(\mathit{p}) \) for each value \( \mathit{p} \) of the protected attribute \( \mathcal{A} \), the window is considered \emph{fair} if, in every block within \( W \), the number of items with attribute value \( \mathit{p} \) lies within the range \([ \lfloor f(\mathit{p}) \times s \rfloor, \lceil f(\mathit{p}) \times s \rceil ]\) for all \( \mathit{p} \). Otherwise, the window is deemed \emph{unfair}. 
The floor and ceiling operations are used because \( f(\mathit{p}) \times s \) may not be an integer, and these bounds ensure the proportion is rounded to the nearest feasible integer values.



Using {\em Constraint 1} listed in Table~\ref{queryconstraints},  this will require every block to have $[\lfloor .3 \times 5\rfloor, \lceil .3 \times 5\rceil]$ of data items of value $\emph{C}$ and value $\emph{A}$ ([1,2] data items) and  $[\lfloor .4 \times 5\rfloor, \lceil .4 \times 5\rceil]$, i.e., $2$ data items with protected attribute value $\emph{H}$. Based on this requirement, the window shown in Figure~\ref{fig:windowblock} is fair.

This definition of fairness is consistent with the notion of \emph{statistical parity} or \emph{demographic parity}~\cite{barocas2019fairness}, appropriately adapted to scenarios where the order of items matters. It can also be viewed as a relaxed variant of \emph{$p$-Fairness}~\cite{wei2022rank}, which enforces the proportionality constraint for every prefix of the ranked list. Finally, depending on the application requirements, one may incorporate an approximation factor~$\epsilon$ in the definition, such that every block within a window satisfies
\[
\left[ \lfloor \epsilon f(\mathit{p}) \times s \rfloor,\ \lceil \epsilon f(\mathit{p}) \times s \rceil \right]
\quad \forall\, \mathit{p}.
\]


\subsection{Proposed Framework}
Given each sliding window and the specified fairness constraint, the framework (Refer to Figure~\ref{fig:framework}) continuously monitors the windows. The sketch is constructed when the first ever window is read. For a new slide after that,  the sketch is updated. After constructing or updating the sketch, it is sent for fairness monitoring. The only other time the sketch is reconstructed is if and only if reordering has been done, as it changes the distribution of items.
\newline Following the definition of \textbf{Fair Sliding Windows} in section ~\ref{sec:dm} if the current window is {\em unfair}, at least one block in the window violates fairness constraints. Since block-level fairness depends on the item arrangement, the \emph{same} set of items may yield either a fair or an unfair window solely based on their permutation. Thus the framework first checks whether it contains a sufficient number of data items so that the items in the window could be reordered to satisfy the constraints. 
This is done by counting the number of instances corresponding to each protected attribute value \( \mathit{p} \), and verifying whether this count meets the required proportion for the entire window. Specifically, if the following condition holds:

\[
\forall \mathit{p} \in \cup_{i=1}^{l} \mathcal{A}_i, \quad \sum_{i=1}^{|W|} \mathbb{1} [\{t_i(A) = \mathit{p}\} \geq (k \cdot \left\lfloor f(\mathit{p}) \cdot s \right\rfloor)]
\]

Basically, for every attribute value \( \mathit{p} \in \cup_{i=1}^{l} \mathcal{A}_i \), it checks if the total number of times \( \mathit{p} \) appears in the window \( W \) (i.e., the number of items \( i \) such that \( t_i(A) = \mathit{p} \)) is at least \( k \cdot \left\lfloor f(\mathit{p}) \cdot s \right\rfloor \), where \( f(\mathit{p}) \) denotes the proportionality of \( \mathit{p} \), \( s \) is block size, and \( k \) is the number of blocks. In that case, the items in the current window can potentially be re-ordered to satisfy the fairness constraint. If the condition holds, then internal reordering can be easily performed within the current window. However, a more challenging scenario arises when the condition does not hold—this is the case we address closely in this work.

In the latter case, the framework pauses and reads an additional \( |\mathcal{X}| \) landmark items. Each landmark item corresponds to a slide, collectively resulting in \( |\mathcal{X}| \) additional sliding windows. The framework then considers both the items in the current window \( W \) and the landmark items \( \mathcal{X} \), and performs reordering. 

The ideal objective is to make both the current window and the \( |\mathcal{X}| \) slid windows satisfy the fairness constraint. However, it is possible that even after incorporating the \( |\mathcal{X}| \) landmark items, the combined set of data items may still lack sufficient instances of certain protected attribute values. 

In such cases, the framework performs a special permutation within \( W \cup \mathcal{X} \) with the goal of maximizing the total number of fair windows among the current window and the \( |\mathcal{X}| \) slid windows. 

After performing the permutation, the monitoring process resumes, and these two steps are repeated as needed until the entire stream has been processed.

\begin{figure}[h]
\centering
\begin{minipage}[t]{0.48\textwidth}
\centering
\begin{tikzpicture}[scale=0.7]
    \node at (-5.4, 0.5) {$B_1$:};
    \scriptsize
    \hspace{-4cm}
    \foreach \i in {1,2,3,4,5} {
        \draw (\i, 0) rectangle (\i + 1, 1);
    }
    \node at (1.5, 0.5) {$[1,\emph{C}]$};
    \node at (2.5, 0.5) {$[2,\emph{C}]$};
    \node at (3.5, 0.5) {$[3,\emph{A}]$};
    \node at (4.5, 0.5) {$[4,\emph{H}]$};
    \node at (5.5, 0.5) {$[5,\emph{H}]$};
\end{tikzpicture}

\vspace{-7.4mm}

\begin{tikzpicture}[scale=0.7]
    \node at (4.6, 1.5) {$B_2$:};
    \scriptsize
    \hspace{3cm}
    \foreach \i in {1,2,3,4,5} {
        \draw (\i, 2) rectangle (\i + 1, 1);
    }
    \node at (1.5, 1.5) {$[6,\emph{C}]$};
    \node at (2.5, 1.5) {$[7,\emph{A}]$};
    \node at (3.5, 1.5) {$[8,\emph{C}]$};
    \node at (4.5, 1.5) {$[9,\emph{H}]$};
    \node at (5.5, 1.5) {$[10,\emph{H}]$};
\end{tikzpicture}

\vspace{3mm}

\begin{tikzpicture}[scale=0.7]
    \node at (-5.4, 0.5) {$B_3$:};
    \scriptsize
    \hspace{-4cm}
    \foreach \i in {1,2,3,4,5} {
        \draw (\i, 0) rectangle (\i + 1, 1);
    }
    \node at (1.5, 0.5) {$[11,\emph{A}]$};
    \node at (2.5, 0.5) {$[12,\emph{A}]$};
    \node at (3.5, 0.5) {$[13,\emph{C}]$};
    \node at (4.5, 0.5) {$[14,\emph{H}]$};
    \node at (5.5, 0.5) {$[15,\emph{H}]$};
\end{tikzpicture}

\vspace{-7.4mm}

\begin{tikzpicture}[scale=0.7]
    \node at (4.6, 1.5) {$\mathcal{X}$:};
    \scriptsize
    \hspace{3cm}
    \foreach \i in {1,2,3,4,5} {
        \draw (\i, 2) rectangle (\i + 1, 1);
    }
    \node at (1.5, 1.5) {$[16,\emph{C}]$};
    \node at (2.5, 1.5) {$[17,\emph{A}]$};
    \node at (3.5, 1.5) {$[18,\emph{A}]$};
    \node at (4.5, 1.5) {$[19,\emph{A}]$};
    \node at (5.5, 1.5) {$[20,\emph{H}]$};
\end{tikzpicture}
\caption{\small Five landmark items $\mathcal{X}$}
\label{fig:landmark}
\end{minipage}%

\hfill
\begin{minipage}[t]{0.48\textwidth}
\centering
\begin{tabular}{l|l}
\hline
\textbf{Symbol} & \textbf{Explanation} \\ \hline
$W$ & Window with width $|W|$ \\ 
$B$ & Block in a window with width $s$ \\ 
$S$ & Sketch \\ 
$k$ & Number of blocks in a window \\ 
$\mathcal{A}$ & Protected attribute with cardinality $\ell$ \\ 
$p$ & One value of $\mathcal{A}$ \\
$\mathcal{X}$ & Landmark items \\ 
$\mathcal{A}^i$ & Protected attribute value of item $i$ \\ \hline
\end{tabular}
\captionof{table}{Key notations}
\label{tab:notation}
\end{minipage}
\vspace{-0.08in}
\end{figure}


\subsection{Problem Definitions}
\begin{problem}\label{prob:readonly}
{\bf Continuous Fairness Monitoring} 
For every sliding window $W$ in the stream and a given fairness constraint, the window is deemed \emph{fair} if each block $B \in W$ satisfies the constraint; otherwise, it is deemed \emph{unfair}.
\end{problem}

The count of a protected attribute value $p$ within a block must lie within the interval $\bigl[\lfloor f(p)\cdot s \rfloor,\; \lceil f(p)\cdot s \rceil \bigr]$
in order for the block to be considered fair with respect to $p$, where $f(p)$ is the fairness constraint for $p$ and $s$ is the block size. For example, if $p = A$, $f(p) = 0.3$, and $s = 5$, then the count of items with $p = A$ in a block must lie in the range $[1,2]$. A block is fair if this condition is satisfied for every protected attribute value. A window is fair if and only if all blocks within the window are fair.

Given the window in Figure~\ref{fig:windowblock} and the {\em Constraint 1} in Table~\ref{queryconstraints}, window $W$ is fair. However, for {\em Constraint 2} in Table~\ref{queryconstraints}, window $W$ is {\em unfair}.

\begin{problem}\label{prob:editable}
 {\bf Continuous Stream Reordering using Landmark Items} 
 Given a window $W$ and additional $|\mathcal{X}|$ landmark items, for a given fairness constraint, identify a reordering for data items in \( W \cup \mathcal{X} \), so that it maximizes the total number of unique fair blocks in  $W$ and the slid windows generated by landmark items.
\end{problem}

Given the \textit{Constraints-2} from Table~\ref{queryconstraints}, the minimum number of $\mathcal{C}$ instances required in window $W$ of Figure~\ref{fig:windowblock} is calculated as $\lfloor 0.5 \times 5 \rfloor \times 3 = 6$. Since $W$ contains only 5 instances of $\emph{C}$, an additional 5 landmark items denoted by $\mathcal{X}$ are read (as shown in Figure~\ref{fig:landmark}). These additional 5 items result in 5 more sliding windows as follows:

\[
W_2: [t_2,t_{16}],\; W_3: [t_3,t_{17}],\; W_4: [t_4,t_{18}],\; W_5:[t_5,t_{19}],\; W_6:[t_6,t_{20}]
\]

$W_1$ from figure ~\ref{fig:landmark} contain blocks 
$B_1=[t_1,\dots,t_5]$, 
$B_2=[t_6,\dots,t_{10}]$, and 
$B_3=[t_{11},\dots,t_{15}]$, 
and let window $W_2$ contain 
$B_1=[t_2,\dots,t_6]$, 
$B_2=[t_7,\dots,t_{11}]$, and 
$B_3=[t_{12},\dots,t_{16}]$.  
All blocks in these two windows are \emph{unique} because each contains at least one item not present in any other block. The blocks of $W_1$ are also \emph{disjoint} since they share no items. We count only unique blocks, as the same set of items may appear in multiple windows (e.g., $[t_6,\dots,t_{10}]$ appears as $B_2$ in $W_1$ and as $B_1$ in $W_6$), and such repetitions should not be double-counted.
\newline Each window is divided into $3$ blocks of size $5$. Thus, there is a unique block starting at each index from $t_1$ through $t_{16}$, giving a total of $16$ unique blocks. These blocks are unique because each block differs from the others by at least one item. 
The ideal goal is to perform reordering among the 20 available items in both $W$ and the landmark items to achieve full fairness whenever feasible. When full fairness is not attainable, the objective is to produce a maximally fair arrangement by maximizing the number of \emph{unique} fair blocks across the current and slid windows.



\section{Continuous Fairness Monitoring}
This section presents our technical solution for continuously and  monitoring fairness constraints over each sliding window \( W \). 
\begin{enumerate}
    \item \textbf{Sketch building:} It first computes the frequency distribution of the protected attribute \( \mathcal{A} \) within the current window \( W \), and stores this information in a lightweight data structure we refer to as the \emph{Forward Sketch}.
    \item \textbf{Fairness check:} Given a fairness constraints, it then leverages the sketch to deem if the window is fair or not.
    \item \textbf{Sketch update:} As the window slides forward, the sketch is updated to reflect the changes, enabling continuous monitoring.
\end{enumerate}

\subsection{Preprocessing}\label{sec:preprocess}
At a high level, sketches~\cite{Indyk06, DatarGIM02, YanXLBCX22} offer compact summaries of data distributions. In our setting, the goal is to capture the distribution of protected attribute values within a sliding window. 

\begin{definition}[\textbf{Forward Sketch}]
A \emph{Forward Sketch} {\tt FSketch} $S$ of width $|W|$ is a data structure associated with a window $W$, where each entry at index $i$ stores a vector of length $(\ell - 1)$ representing the cumulative frequencies of $(\ell - 1)$ protected attribute values from the beginning of the window up to position $i$. The $j$-th entry of the vector corresponds to the $j$-th value of the protected attribute $\mathcal{A}$. Since the protected attribute $\mathcal{A}$ has $\ell$ unique values, the frequency of the remaining protected attribute value at index $i$ can be inferred by subtracting the sum of the stored frequencies from $i$, the total number of items considered up to that point.
\end{definition}
    
   Consider the sketch $S$ in Figure~\ref{fig:sketchcreation} that shows blocks $B_1$ and $B_2$ of the running example. The first bit of the vector is assigned to store the cumulative frequency of $\emph{C}$ and the second one is that of for $\emph{A}$. As an example, the $4$-th entry of $S$ contains $[2,1]$, denoting a total frequency of $2$ of $\emph{C}$ and $1$ of $\emph{A}$ from the beginning till the $4$-th position of the window. Since $\mathcal{A}$ has three distinct values, the frequency of $\emph{H}$ could be inferred by subtracting each $i$ from the cumulative frequency of $\emph{C}$ and $\emph{A}$ at position $i$. As an example, the cumulative frequency of $\emph{H}$ from the beginning upto position $i$ is therefore $4-(2+1)=1$.

\subsubsection{Sketch Construction Algorithm}
Each entry $i$ of the sketch $S$ stores a vector of length $(\ell - 1)$, representing the cumulative counts of the first $(\ell - 1)$ protected attribute values from the start of the window up to position $i$. For $i > 1$, the sketch at index $i$ copies the counts from the previous index ($i - 1$), and then increments the count corresponding to the $j$-th protected attribute value $\mathit{p}$ if the item $t_i$ has $\mathcal{A}^i = \mathit{p}$. If the encountered value is not stored explicitly in the vector, it does not do anything. This construction ensures that each sketch entry maintains a cumulative sum of the $\ell-1$ protected attribute values up to that point in the window. From there, the value of the protected attribute that is not stored could always be calculated. Algorithm~\ref{alg:sketch} has the pseudocode.

\begin{algorithm}[!htbp]
\caption{Forward Sketch Construct - {\tt FSketch} }\label{alg:sketch}
\begin{algorithmic}[1]
\Require Window $W = \{t_1, t_2, \ldots, t_{|W|}\}$ of size $|W|$
\Require Set of protected attribute values $\mathcal{A} = \{p_1, p_2, \ldots, p_\ell\}$
\State Initialize sketch $S$ as list of $|W|$ vectors, each of length $(\ell - 1)$, with all entries set to $0$
\For{$i = 1$ to $|W|$}
    \If{$i > 1$}
        \State $S[i] \gets S[i-1]$ \Comment{Copy counts from previous entry}
    \EndIf
    \State Let $\mathit{p} \gets \mathcal{A}^{t_i}$ \Comment{Protected attribute value of item $t_i$}
    \For{$j = 1$ to $(\ell - 1)$}
        \If{$\mathit{p} = p_j$}
            \State $S[i][j] \gets S[i][j] + 1$
            \State \textbf{break}
        \EndIf
    \EndFor
    \Comment{If $\mathit{p} = p_\ell$, no update is made}
\EndFor
\State \Return $S$
\end{algorithmic}
\end{algorithm}

\begin{theorem}
    Algorithm {\tt Construct-Sketch} (Algorithm~\ref{alg:sketch}) takes $\mathcal{O}(|W|\times \ell)$ to run and takes $\mathcal{O}(|W| \times \ell)$ space.
\end{theorem}

\begin{proof}
From Algorithm~\ref{alg:sketch}, it is evident that each entry in the sketch requires $\mathcal{O}(\ell)$ time to compute and $\mathcal{O}(\ell)$ space to store, as it maintains information for $\ell - 1$ protected attribute values. Since the outer loop iterates over the entire window of size $|W|$, the overall time and space complexity of the algorithm is $\mathcal{O}(|W| \times \ell)$.
\end{proof}

\subsubsection{Sketch Update}
During the first time of creation, the sketch initializes by tracking $(\ell - 1)$ protected attribute values, starting from the first index of the first window. This ensures that the earliest data items are captured first. The update ensures that the sketch accurately represents the most recent window of data. When window $W$ is slid over, {\em the sketch shifts all entries one position to the left}. The last entry in the sketch is then updated considering the last but one entry and with the protected attribute value of the newly added item. This way, the sketch remains synchronized with the current window and continues to provide an up-to-date, compact summary of the protected attributes over time.

Consider the sketch in Figure~\ref{fig:sketchcreation} again. When the window slides over by one item, the $11$-th item comes in. All entries in the previous sketch (refer to Figure~\ref{fig:fwdsketch}) are now moved one position to the left, and the last entry of the sketch is now updated.

\begin{lemma}
The sketch takes $O(\ell)$ time to update.
\end{lemma}
\begin{proof}
It takes only a constant time to remove the first element of the sketch. A new vector is appended at the end, whose amortized cost is $O(\ell)$ time.
\end{proof}

\begin{figure}[!htbp]
        \hspace{0.4cm}
    \begin{tikzpicture}[scale=0.8]
        \node at (-5.4, 0.5) {$S$:};
        \scriptsize
        \hspace{-4.66cm}
        \foreach \i in {1,2,3,4,5,6,7,8,9,10} {
             \draw (\i, 0) rectangle (\i + 1, 1);  
        }
            \node at (1 + 0.5, 0.5) {$[1,0]$};
            \node at (2 + 0.5, 0.5) {$[2,0]$};
            \node at (3 + 0.5, 0.5) {$[2,1]$};
            \node at (4 + 0.5, 0.5) {$[2,1]$};
            \node at (5 + 0.5, 0.5) {$\textbf{[2,1]}$};
            \node at (6 + 0.5, 0.5) {$[3,1]$};
            \node at (7 + 0.5, 0.5) {$[3,2]$};
            \node at (8 + 0.5, 0.5) {$[4,2]$};
            \node at (9 + 0.5, 0.5) {$[4,2]$};
            \node at (10 + 0.5, 0.5) {$\textbf{[4,2]}$};
         \end{tikzpicture}
    \caption{\small Sketch of blocks $B_1$ and $B_2$ in Figure~\ref{fig:windowblock}}
\label{fig:sketchcreation}    
\end{figure}

\begin{figure}[!htbp]
\centering
\scriptsize
\begin{tikzpicture}[scale=0.8]

    \node at (-0.5, 0.5) {$W$:};
    \scriptsize
    \hspace{-1cm}
    \foreach \i in {1,2,3,4,5,6,7,8,9,10} {
         \draw (\i, 0) rectangle (\i + 1, 1);  
    }
        \node at (1.5, 0.5) {$[2,\mathcal{C}]$};
        \node at (2.5, 0.5) {$[3,\mathcal{A}]$};
        \node at (3.5, 0.5) {$[4,\mathcal{H}]$};
        \node at (4.5, 0.5) {$[5,\mathcal{H}]$};
        \node at (5.5, 0.5) {$[6,\mathcal{C}]$};
        \node at (6.5, 0.5) {$[7,\mathcal{A}]$};
        \node at (7.5, 0.5) {$[8,\mathcal{C}]$};
        \node at (8.5, 0.5) {$[9,\mathcal{A}]$};
        \node at (9.5, 0.5) {$[10,\mathcal{H}]$};
        \node at (10.5, 0.5) {$[11,\mathcal{A}]$};

    \hspace{1cm}
    \node at (-0.5, -1) {$S$:};
     \scriptsize
    \hspace{-1cm}
    \foreach \i in {1,2,3,4,5,6,7,8,9,10} {
         \draw (\i, -1.5) rectangle (\i + 1, -0.5);  
    }
        \node at (1.5, -1) {$[2,0]$};
        \node at (2.5, -1) {$[2,1]$};
        \node at (3.5, -1) {$[2,1]$};
        \node at (4.5, -1) {$[2,1]$};
        \node at (5.5, -1) {$[3,1]$};
        \node at (6.5, -1) {$[3,2]$};
        \node at (7.5, -1) {$[4,2]$};
        \node at (8.5, -1) {$[4,2]$};
        \node at (9.5, -1) {$[4,2]$};
        \node at (10.5, -1) {$[4,3]$};
\end{tikzpicture}
\caption{\small Sketch maintenance with one item slide}
\label{fig:fwdsketch}
\end{figure}

\subsection{Fairness Checking}\label{sec:qp}
Next, we discuss Algorithm {\tt Monitor-BFair} that checks whether a window is fair or not. The innovation of the algorithm is that it only needs to check at most the number of blocks $k$ number of entries in the current sketch $S$, and returns the correct answer always. In some cases, it is able to terminate early, especially when it already has found violation of the fairness constraints. Specifically it only checks those entries of the current sketch $S$ that represents the last entry of each block. 

To evaluate the first block of size $s$, Algorithm {\tt Monitor-BFair} first examines the $s$-th entry of the sketch $S$, which contains cumulative counts for the first $\ell - 1$ protected attribute values. The count for the final (implicit) attribute is computed by subtracting the sum of the stored counts from item number. The fairness constraint is verified by checking whether all of these $\ell - 1$ counts—and the inferred count of the $\ell$-th attribute—satisfy the specified requirements.

For each subsequent block, the algorithm computes the difference between two sketch vectors : specifically, the last entry of the previous block and that of the current block. This difference yields the frequency of each protected attribute value within the current block. The process repeats for all blocks, continuing until either a block violates the fairness constraints—causing the algorithm to terminate early and return {\em No}—or all blocks pass the checks, in which case {\tt Monitor-BFair} returns {\em Yes}. Algorithm~\ref{alg:qp} presents the pseudocode.

Consider Figure~\ref{fig:sketchcreation} and Constraints 1 from Table~\ref{queryconstraints}. The window $W$ contains 10 elements, which {\tt Monitor-BFair} partitions into two blocks, ${B_1, B_2}$, each of size $s = 5$.
The algorithm starts by examining the sketch at the end of the first block, i.e., at index 5, where the sketch holds the cumulative vector $[2, 1]$. This indicates that within $B_1$, there are 2 instances of $\emph{C}$ and 1 instance of $\emph{A}$, implying the remaining 2 items correspond to $\emph{H}$. The block satisfies the first fairness constraint. Next, the algorithm proceeds to the end of the second block, at index 10. It computes the frequency vector for $B_2$ by subtracting the sketch at index 5 from that at index 10. The resulting vector reveals that $B_2$ contains 2 instances of $\emph{C}$, 1 instance of $\emph{A}$, and 2 instances of $\emph{H}$. This also satisfies the specified fairness requirements. This process continues for all three blocks. Finally, algorithm concludes that the window to be {\em fair}.

On the other hand, consider Figure~\ref{fig:sketchcreation} and Constraints 2 from Table~\ref{queryconstraints}. Its first two blocks would be deemed fair, but the third block will not, hence, the window is {\em unfair}.

\begin{algorithm}
\caption{\texttt{Monitor-BFair}:Fairness Checking Algorithm}\label{alg:qp}
\begin{algorithmic}[1]
\Require Sketch $S$ with $|W|$ entries indexed from $1$ to $|W|$;
\Require Block size $s$, number of blocks $k$;
\Require Fairness proportions $f(p)$ for each $p \in \{1, \dots, \ell\}$
\Ensure \textbf{Fair} if all blocks satisfy fairness constraints; \textbf{Unfair} otherwise
For{$b \gets 1$ to $k$}
    \State $curr \gets b \cdot s$
    
    \If{$b = 1$}
        \State $countVec \gets S[curr]$
    \Else
        \State $prev \gets (b - 1) \cdot s$
        \State $countVec \gets S[curr] - S[prev]$
    \EndIf

    \State $total \gets s$
    \State $lastCount \gets total - \sum_{j=1}^{\ell - 1} countVec[j]$
    \State Construct $fullVec[1 \dots \ell] \gets (countVec[1 \dots \ell - 1], lastCount)$

    \For{$p \gets 1$ to $\ell$}
        \State $minReq \gets \lfloor f(p) \cdot s \rfloor$
        \State $maxReq \gets \lceil f(p) \cdot s \rceil$
        \If{$fullVec[p] < minReq$ \textbf{or} $fullVec[p] > maxReq$}
            \State \Return \textbf{Unfair}
        \EndIf
    \EndFor
\State \Return \textbf{Fair}
\end{algorithmic}
\end{algorithm}

\begin{lemma}
    Worst case running time of {\tt Monitor-BFair} is $\mathcal{O}(k \times \ell)$
\end{lemma}
\begin{proof}
    At the worst case,  {\tt Monitor-BFair} needs to process all $k$ blocks, each requiring $\mathcal{O}(\ell)$ time. Therefore, it overall will take $\mathcal{O}(k \times \ell)$ time.
\end{proof}

\begin{lemma}
    Best case running time of {\tt Monitor-BFair} is ${\Omega}(\ell)$
\end{lemma}
\begin{proof}
    At the best case,  {\tt Monitor-BFair} finds fairness violation in the first block itself, requiring  ${\Omega}(\ell)$ time to declare the block {\em unfair} and terminate.
\end{proof}

\begin{lemma}
  Algorithm {\tt Monitor-BFair} is optimal.
\end{lemma}
We refer to the technical report in \cite{git} for the proof.

\section{Continuous Stream Reordering } \label{sec:reorder}
Consider fairness \textbf{Constraints 2} again, there are three protected attribute values: \( C, A, H \). Two distinct fairness requirements are specified:

\noindent\textbf{Case 1:} The requirement is \([2, 1, 2]\), meaning each block must contain 2 elements with value \( \emph{C}\), 1 with \( \emph{A} \), and 2 with \( \emph{H} \). \\
\noindent\textbf{Case 2:} The requirement is \([3, 1, 1]\).

A block is considered fair if it satisfies either of these criteria. Since the original block \( B_3 \) does not meet either requirement, 5 additional landmark items are read (Figure~\ref{fig:landmark}). The updated input stream, including these landmark items, now consists of 6 elements with value \( C \), 7 with \( A \), and 7 with \( H \). We propose an algorithm, {\tt BFair-ReOrder} , which takes these 20 elements and rearranges them to maximize the number of uniquely fair blocks.

Let $n=|W|+|\mathcal{X}|$ and \( InStm = [ib_1, ib_2, \ldots, ib_n] \), be the stream under consideration with $|W|$ regular and $|\mathcal{X}|$ landmark items. The objective is to permute the elements into a new sequence \( OutStm = [ob_1, ob_2, \ldots, ob_n] \) to maximize the number of unique fair blocks. Denote by \( fb(Stm) \) the number of unique fair blocks in stream \( Stm \). Then the problem can be formulated to maximize
$OutStm = \arg\max_{Stm} fb(Stm)$.

\subsection{Algorithm {\tt BFair-ReOrder} }
Algorithm {\tt BFair-ReOrder}  has an outer loop that takes each valid combination of protected attribute counts and runs a subroutine (Subroutine {\tt MaxReOrder}) to reorder and create the maximum number of unique fair blocks with that. For {\bf Constraints~2}, there are two such combinations for \([C, A, H]\): \([2,1,2]\) and \([3,1,1]\), and it repeats the subroutine for each and outputs the one that maximizes the number of fair unique blocks at the end. 

We begin by introducing key concepts used in the design of the Subroutine {\tt MaxReOrder}.

\begin{definition}[Isomorphic Blocks and Isomorphic Block Count]
A set of disjoint blocks is said to be \emph{isomorphic} if each block contains the same ordered sequence of protected attribute values. The total number of such blocks in a stream is called the \emph{Isomorphic Block Count (IBC)}. Under a given fairness constraint, if one isomorphic block satisfies the fairness condition, then all blocks in the set are considered fair.
\end{definition}
Assume the input stream consists of $\ell$ distinct attribute values. Let \( v_j \geq 1 \) denote the number of elements with value \( p \in \cup_{i=1}^{\ell} \mathcal{A}_i \) required by the fairness constraint, such that
$
\sum_{j=1}^{\ell} v_j = s.
$ 
Let \( V_j \) be the total number of elements of value \( p \) in the input stream. Then, the number of \emph{isomorphic blocks} (IBC) that satisfy the given fairness requirement can be computed as:
\[
\text{IBC} = \min\left\{ \text{IBC}_j \mid 1 \leq j \leq \ell \right\}, \quad \text{where } \text{IBC}_j = \left\lfloor \frac{V_j}{v_j} \right\rfloor.
\]

For example, in \textbf{constraints 2}, \noindent\textbf{Case 1:} \[
\left\lfloor \frac{6}{2} \right\rfloor = 3, \quad \left\lfloor \frac{7}{1} \right\rfloor = 7, \quad \left\lfloor \frac{7}{2} \right\rfloor = 3.
\]
Thus, the total number of fair isomorphic blocks is:
\[
\text{IBC} = \min(3, 7, 3) = 3.
\]

\noindent For \textbf{Case 2:}, $\text{IBC} = \min(2, 7, 7) = 2$

\begin{definition}[Isomorphic Stream and Extended Isomorphic Stream]
A stream is called an \emph{isomorphic stream} if it is entirely composed of isomorphic blocks—that is, each block follows the same fixed composition pattern. If such a stream is extended by a prefix of this block pattern, the resulting stream is called an \emph{extended isomorphic stream}. This additional segment is referred to as the \emph{extended prefix (EP)}, and its length is called the \emph{extended prefix length (EPL)}.
\end{definition}

Given a stream of length \( n \) and a fairness requirement specifying a block size \( s \), 
if \( IBC \cdot s = n \), the stream can be fully partitioned into \( IBC \) isomorphic blocks, or it can be reordered into an isomorphic stream.

If \( IBC \cdot s < n \), the remaining \( n - IBC \cdot s \) items cannot form a complete isomorphic block. These leftover elements may form an extended prefix \emph{EP}. 
The length of the extended prefix is given by:
\[
EPL = \sum_{i=1}^{\ell} EP_i, \quad \text{where } EP_i = \min\left(v_i,\ V_i - IBC \cdot v_i\right)
\]
Here,
  \( \ell \) is the number of distinct values (e.g., demographic groups) in the protected attribute;
  \( v_i \) is the required count of group \( i \) per fair block (from the fairness requirement);
  \( V_i \) is the total number of elements from group \( i \) in the entire stream;
  \( EP_i \) is the number of group \( i \) elements assigned to the extended prefix;
  \( IBC \) is the number of complete isomorphic blocks.

\noindent Note: \( EPL < s \), since the extended prefix cannot complete a full block on its own.

Consider the fairness requirement of \([2, 1, 2]\) for  \( C \), \( A \), and \( H \), where $C = 6,  A = 7, H = 7$.
  
The block size is \( s = 5 \), and since \( IBC = 3 \), we calculate:
$EP_1 = \min(2,\ 6 - 3 \cdot 2) = 0, EP_2 = \min(1,\ 7 - 3 \cdot 1) = 1, 
EP_3 = \min(2,\ 7 - 3 \cdot 2) = 1$, which gives: $EPL = EP_1 + EP_2 + EP_3 = 0 + 1 + 1 = 2$. So the extended prefix is \( EP = [A, H] \) (any permutation is valid). To complete the isomorphic block pattern, use the remaining counts \( v_i - EP_i \):
\[
\begin{aligned}
C: &\quad 2 - 0 = 2 \quad \Rightarrow [C, C] \\
A: &\quad 1 - 1 = 0 \quad \Rightarrow [] \\
H: &\quad 2 - 1 = 1 \quad \Rightarrow [H]
\end{aligned}
\]
One valid isomorphic block pattern is: \( [A, H, C, C, H] \). We can now construct the extended isomorphic stream:
\[
[A, H, C, C, H, \; A, H, C, C, H, \; A, H, C, C, H, \; A, H]
\]
This consists of three full isomorphic blocks and a prefix of length 2, achieving optimal block-wise fairness. The 3 leftover $A$s can be appended at the end to complete the stream.

In some situations, more than one cases of the fairness constraint maybe involved to make the stream maximally fair. For example: Consider the distribution of items in the stream as \([C, A, H]\): \([7,8,5]\). We obtain \( \text{IBC}_{Case 1} = 2 \) for required pattern $[2,1,2]$. The remaining $[3,6,1]$ items can be arranged following Case 2's pattern $[3,1,1]$ which also satisfies the fairness constraint.
\newline After getting the \( \text{IBC} = 2 \) for required pattern $[2,1,2]$  where $C = 7,  A = 8, H = 5$ we build the $EP$: $EP_1 = \text{min}(2, 7 - 2\cdot2) = 2$, $EP_2 = \text{min}(1, 8 - 2\cdot1) = 1$ and $EP_3 = \text{min}(2, 5 - 2\cdot2) = 1$, which gives $EPL = 2 + 1 + 1 = 4$. So the $EP = [C,C,H,A]$ (any permutation is valid for now). 
\newline Now with the remaining items $R = [3,6,1]$, we compute $\text{IBC}_R = min(1, 6, 1) = 1$ for the Case 2 pattern $[3,1,1]$. We build $EP_R$: $EP_{R1} = \text{min}(3, 3 - 1\cdot3) = 0$, $EP_{R2} = \text{min}(1, 6 - 1\cdot1) = 1$ and $EP_{R3} = \text{min}(1, 1 - 1\cdot1) = 0$. So the $EPL_R = 0+1+0 = 1$ and $EP_R = [A]$. 
\newline Therefore, all the blocks must have prefix pattern $[A]$. For the first three blocks that can be made isomorphic, the prefix follows the pattern $[A,C,C,H]$ (Only A's position is fixed, for the rest any permutation is valid again). Thus the first 2 isomorphic blocks following Case 1 are: \( [A, C, C, H, H], [A, C, C, H, H]\). The third block following Case 2 is: \( [A, C, C, H, C]\). Thus the complete isomorphic stream is:
\[
[A, C, C, H, H, A, C, C, H, H, A, C, C, H, C, A]
\]
The 4 leftover $A$s can be appended to complete the stream.

To summarize, Subroutine {\tt MaxReOrder}(see Subroutine~\ref{alg:isomorphic1}) begins by computing \( IBC \), the number of isomorphic blocks that can satisfy the given fairness requirement. If no block can satisfy the constraint, the algorithm simply returns the original stream.

If \( IBC \cdot s = n  \), where \( s \) is the block size and \( n \) is the total stream length, then the entire output stream can be constructed using exactly \( IBC \) isomorphic blocks, forming an \emph{isomorphic stream}. In this case, the algorithm returns the built isomorphic stream directly.

If \( IBC \cdot s < n \), the remaining elements—those that do not fit into the full isomorphic blocks—can be used to form an \emph{extended prefix} \( EP \). The extended stream is constructed by concatenating the \( IBC \) isomorphic blocks with this extended prefix. This stream contains the maximum number of fair blocks possible under the given constraints (see Lemma~\ref{lemma:isomorphicstream}). Any leftover elements that are not part of the isomorphic blocks or the extended prefix are appended to complete the stream. 





\begin{algorithm}[!htbp]
\caption{Subroutine {\tt MaxReOrder}}
\label{alg:isomorphic1}
\begin{algorithmic}[1]
\State \textbf{Input:} Input stream \( InStm \) of length \( |W| + \mathcal{X} \), fairness constraint \( [v_1, \dots, v_\ell] \)
\State \textbf{Output:} Reordered stream \( OutStm \)

\State Compute the number of fair isomorphic blocks \( IBC \) \label{iso:calculateIB}
\If{\( IBC < 1 \)} \label{iso:nofairblock}
    \State \Return \( InStm \)  
\EndIf \label{iso:nofairblockend}

\If{\( IBC \cdot s = n \)} \label{iso:exactfairblocks}
    \State Construct \( IBC \) fair isomorphic blocks based on the given fairness constraint
    \State \( OutStm \gets \) isomorphic stream formed by the \( IBC \) fair isomorphic blocks
    \State \Return \( OutStm \)
\EndIf \label{iso:exactfairblocksend}

\State Build the extended prefix \( EP \)
\If { \( EP \) is not part of any new fair block 
    \State Complete the Isomorphic block pattern using \( v_i - EP_i \)
    \State Construct an extended stream by concatenating the \( IBC \) fair isomorphic blocks and \( EP \)
    \State \( OutStm \gets \) extended stream + remaining elements 
    \State \Return \( OutStm \) \label{iso:returnisopatterned}    
\Else
    \State Calculate \(IBC_R\) based on the remaining part and the new constraint
    \State Build the extended prefix \(EP_R\) based on the remaining part and the new constraint
    \State Complete the \( IBC, IBC_R \) Isomorphic block patterns
    \State \( OutStm \gets  \) \( IBC+IBC_R \) fair isomorphic blocks + extended prefix \(EP_R\)+final remaining part
    \State \Return \( OutStm \) \label{iso:returnmultipleconstraints}
\EndIf}  

\end{algorithmic}
\end{algorithm}

\subsection{Optimality and Running Time Analysis}
\begin{lemma}[Optimality of Isomorphic and Extended Isomorphic Streams] \label{lemma:isomorphicstream}
\end{lemma}
\begin{proof}
\noindent Let the stream have length \( n \) and block size \( s \). The number of possible unique blocks of size \( s \) in the stream is \( n - s + 1 \). We say a stream is \emph{optimal} if all these blocks are \emph{fair}—that is, it contains the maximum possible number of unique fair blocks.

\noindent An \emph{isomorphic stream} consists of \( \text{IBC} = \frac{n}{s} \) disjoint, identical fair blocks of size \( s \). Because each overlapping block (i.e., any block of \( s \) consecutive elements) is composed of suffixes and prefixes of these identical fair blocks, every such block also satisfies the fairness condition. Therefore, all \( n - s + 1 \) unique blocks are fair, and the stream achieves optimality.

Imagine two consecutive isomorphic blocks:
\[IB1 = [<1,A>, <2,C>, <3,C>, <4,H>, <5,H>],\] \[IB2 = [<6,A>, <7,C>, <8,C>, <9,H>, <10,H>]\] The overlapping block 
$U=[<3,C>, <4,H>, <5,H>, <6,A>, <7,C>]$ is exactly the suffix of IB1 plus the prefix of IB2. Since both IB1 and IB2 satisfy the required fairness counts, any block composed of the suffixes and prefixes of these isomorphic blocks-including 
U must also be fair.

\noindent An \emph{extended isomorphic stream} includes a carefully constructed prefix \( EP \) appending the isomorphic blocks. This prefix is designed such that any overlapping block between \( EP \) and the last isomorphic block is fair. Since all blocks within the isomorphic part are also fair, the entire stream contains \( n - s + 1 \) fair blocks. Thus, the extended isomorphic stream is also optimal.

\end{proof}

\vspace{-0.12in}

\begin{theorem}[Algorithm {\tt BFair-ReOrder}  is optimal] \label{theorem:optimal}
\end{theorem}

\begin{proof}
If \( IBC \cdot s = n \), the stream is an isomorphic stream, and the result follows from Lemma~\ref{lemma:isomorphicstream}.

If \( IBC \cdot s < n \), the extended stream includes \( IBC \cdot s + EPL \) elements, which contribute \( IBC \cdot s + EPL -s  + 1 =(IBC - 1) \cdot s + EPL + 1\) fair blocks. Based on the construction method, the remaining elements cannot complete another fair block or contribute to overlaps of any fair block. Thus, this is the maximum number of fair blocks possible. Subroutine {\tt MaxReOrder} finds that optimality for each valid combination of protected attribute values and Algorithm {\tt BFair-ReOrder}  checks this for all possible combinations and outputs the one that maximizes it.
\end{proof}

\noindent {\bf Running time.}  Subroutine {\tt MaxReOrder} takes at most  \( \mathcal{O}(n) \) ($n= |W|+|\mathcal{X}|$) time. At the worst case, {\tt MaxReOrder} needs to be run $2^{\ell}$ times. Therefore, Algorithm {\tt BFair-ReOrder}  takes In \( \mathcal{O}(n \times 2^{\ell}) \) time. The space complexity is dominated by queue size, $\mathcal{O}(n)$.

\section{Experimental Evaluations}
We present experimental analyses that focus on three aspects: (a) the effectiveness of the fairness model in ensuring group fairness at the block level, (b) the effectiveness of fairness checking and reordering solutions, and (c) the effectiveness and cost of the pre-processing framework. Code and data are available at ~\cite{git}.

\subsection{Experimental Setup}
\smallskip  \noindent{\bf Streaming environment.} The streaming environment is simulated using Confluent Kafka Python (CKP), the official Python client for Apache Kafka. A single-broker exposes an external listener to receive data from a host-based producer, which streams JSON-formatted dataset records to a Kafka topic at a controlled rate.

\noindent The CKP-based consumer polls the broker, maintains an in-memory buffer of the latest $|\emph{W}|$ messages, and forms a sliding window $\emph{W}$ for preprocessing and fairness checking. 
 
\smallskip  \noindent{\bf Hardware and Software.} All experiments are conducted on an HP-Omen 16-n0xxx (Windows 11 Home v10.0.26100, Python 3.9.13) with an 8-core AMD Ryzen 7 6800H CPU and 16 GB RAM. A single-node Kafka cluster (v3.7.0) and ZooKeeper (both from Confluent Platform 7.8.0) run in Docker (v4.41.2), simulating the streaming environment. Results are averaged over 10 runs.

\smallskip  \noindent {\bf Datasets.}
The experiments are conducted using $4$ real world datasets (Table \ref{tab:dataset} has further details).\\
\noindent \textbf{Hospital}~\cite{diagnostics12020241}: Contains over 10,000 patient admissions from a tertiary care hospital in India (April 2017–March 2019), sorted by admission date. The “AGE” attribute was quantile-binned into five bins.\\
\noindent \textbf{Movies}~\cite{movielens}: Includes 33M+ ratings from 200K+ users on 87,585 movies. Genres were mapped to 2 and 3 “moods” (dark/light/neutral) categories, using joined movie and rating tables.\\
\noindent \textbf{Stocks}~\cite{oleh_onyshchak_2020}: Tracks daily NASDAQ prices up to April 2020. Apple Inc.'s price changes were computed as percentage differences between open and close, binned into five equal-sized bins. Trading volume was binned into 2 and 3 categories.\\
\noindent  \textbf{Tweets} \cite{twittersent}: Contains 16M tweets (April–May 2009). Sentiment analysis\cite{loria2018textblob} was applied to assign one of five polarity-based sentiment categories.

\noindent {\bf Baselines.}
We note that {\em there is no related work that could be adapted to solve the two problems studied in the paper (see Section~\ref{sec:related} for further details)}. Therefore, we design our own baselines. \\
\noindent {\bf Backward Sketch} or \textit{BSketch}. This is designed to compare against the window preprocessing technique discussed in Section~\ref{sec:preprocess} referred to as  {\tt FSketch}. We call this \textit{BSketch} as this starts constructing the sketch backward. Basically, at index $i$ it captures the cumulative counts of the $\ell-1$ protected attribute values starting $i$-th position till the end of the window. Naturally, for every new window, the sketch must be reconstructed from scratch because all items in sketch will need fresh update (not just shifts unlike forward sketch).\\
\noindent {\bf BSketchQP} For monitoring continuous fairness that uses \textit{BSketch} for preprocessing. This is designed to compare against {\tt Monitor-BFair} proposed in Section~\ref{sec:qp}.\\
\noindent {\bf BruteForce} that enumerates all possible reordering and selects the one that maximizes the number of unique fair blocks to compare against {\tt BFair-ReOrder} discussed in Section~\ref{sec:reorder}. Naturally, \textit{BruteForce} is computationally intensive and does not scale at all. \\
\noindent {\bf Measures.} 
\noindent  For fairness analysis, we measure the percentage of fair blocks produced by our algorithm relative to the original stream. Then, we present a small qualitative study that compare block level fairness with window level fairness  based on user satisfaction.  We also evaluate the optimality of {\tt BFair-ReOrder}. We measure the average fairness checking time, 90\% tail latency (i.e., the time within which 90\% of the check is complete, while 10\% take longer), throughput (number of fairness constraints answered per unit time), and memory consumption. For preprocessing analysis, we report sketch construction and update time, memory consumption, as well as 90\% tail latency of preprocessing. We also present running time analysis of each component. \\
 \noindent{\bf Fairness constraints and Parameters.} We vary window size $|W|$, block size $s$, cardinality of the protected attribute $\ell$, landmark size $|\mathcal{X}|$, as appropriate. Fairness constraints are generated from the proportion of protected attribute values present in the respective datasets. The protected attributes used and their respective $\ell$ are discussed in Table~\ref{tab:dataset}; one protected attribute at a time. The default parameter values are $|W|$=1000, $s$=25, 
$|\mathcal{X}|$=100 and $\ell=5$.

\begin{table}[!htbp]
    \centering
    \begin{tabular}{|p{0.15\textwidth}|p{0.2\textwidth}|p{0.07\textwidth}|}\hline
\textbf{ Datasets used}&\textbf{Protected attribute($\ell$)}& \textbf{Records}\\\hline
         Hospital & Gender(2),Outcome(3),Age(5)& 10,101 \\\hline
         Twitter & Sentiments(5) & 16,000,000 \\\hline
         MovieLens & Ratings(5), Moods(2,3) & 33,832,161 \\\hline
         Stock & \% price change(5), volume of trade(2,3) & 9,908 \\ \hline
    \end{tabular}
    \caption{Datasets \& statistics}
    \label{tab:dataset}
\end{table}
\vspace{-0.2in}

\noindent{\bf Summary of results.} Our experimental analysis reveals several key findings. First, our qualitative study demonstrates that the users unanimously prefer our proposed fairness model, as opposed to enabling group fairness only on  the overall window. Second, the empirical results are consistent with our theoretical guarantees for the proposed algorithms used in preprocessing, reordering, and fairness checking. Third, our framework outperforms all designed baselines by 2--3 orders of magnitude in terms of efficiency. Fourth, it effectively enforces fairness at the block level, substantially increasing the percentage of fair blocks when {\tt BFair-ReOrder} is applied. Additionally, the cost of fairness monitoring remains minimal in both time and memory: {\tt Monitor-BFair} answers queries in fractions of a millisecond, and {\tt BFair-ReOrder} scales linearly with the input stream size. Overall, the framework is accurate and scalable across a wide range of parameters.

\subsection{Fairness Analyses \& Qualitative Study}

\begin{table}[!htbp]
    \centering
    \setlength{\tabcolsep}{-2.5pt} 
    \renewcommand{\arraystretch}{1.1} 
    \begin{tabular}{p{5cm} >{\centering\arraybackslash}p{1.5cm} >{\centering\arraybackslash}p{1.5cm} >{\centering\arraybackslash}p{1.5cm} >{\centering\arraybackslash}p{1.5cm}}
    \toprule
         \textbf{Datasets}&  \textbf{Parameter setting}&  \textbf{Before Reorder}& \textbf{After Reorder}\\
    \midrule
         &  $|W|$ = 200&  0\%& 62\%\\
         Hospital ($s$=25, $|\mathcal{X}|$ = 100, $l$ = 5)&  $|W|$ = 1000&  8\%& 95\%\\
         &  $|W|$ = 2000&  1\%& 93\%\\
         &  $s$ = 25&  0\%& 32\%\\
         Tweets ($|W|$=1000, $|\mathcal{X}|$ = 500, $l$ = 5)&  $s$ = 100&  0\%& 87\%\\
         &  $s$ = 250&  0\%& 96\%\\
         &  $l$ = 2&  2\%& 98\%\\
         Movies ($|W|$=1000, $s$ = 250, $|\mathcal{X}|$ = 500)&  $l$ = 3&  0\%& 99\%\\
         &  $l$ = 5&  0\%& 21\%\\ 
    \bottomrule
    \end{tabular}
    \caption{\small Fairness achieved by {\tt BFair-ReOrder}}
    \label{tab:FairnessTable}
    \vspace{-0.2in}
\end{table}


\subsubsection{Fairness Analyses}
We compare our proposed reordering algorithm {\tt BFair-ReOrder}  with {\tt BruteForce} to empirically evaluate our theoretical claim. We observe that {\tt BFair-ReOrder}  returns the same number of unique fair blocks as that of  {\tt BruteForce} corroborating its optimality. Additional results could be found in the technical report in ~\cite{git}.

We vary window size $|W|$, block size $s$, and cardinality $\ell$ of different datasets and present the percentage of unique blocks that become fair after {\tt BFair-ReOrder}  is applied on the original stream. Table~\ref{tab:FairnessTable} presents a subset of representative results. A general observation is, {\tt BFair-ReOrder}  significantly improves the fairness of the data stream.
In general, with increasing window size $|W|$, as well as block size $s$, fairness substantially improves, which is intuitive, because a larger window or block size increases the likelihood of more number of blocks satisfying fairness for the same constraints. On the other hand, with increasing cardinality, the group fairness requirement becomes more complex, leading to less percentage of fair blocks. 

\begin{figure}[!htbp]
\begin{center}
\includegraphics[width=0.33\textwidth]{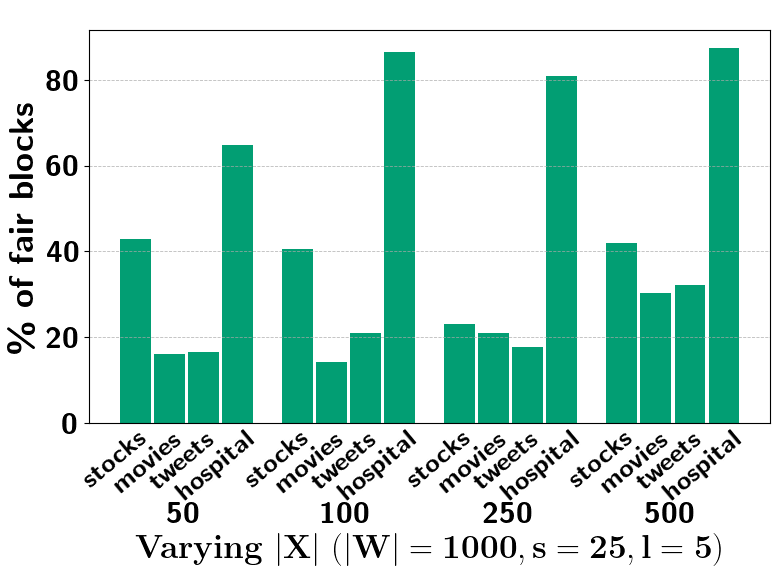}
\end{center}
\vspace{-0.2in}
\caption{\small Effect of landmark on fairness}\label{fig:land}
\end{figure}

In Figure~\ref{fig:land}, we vary $|\mathcal{X}|$ and measure the percentage of unique fair blocks after {\tt BFair-ReOrder}  is applied. The results indicate that with increasing $|\mathcal{X}|$, the percentage of unique fair blocks increases - this is intuitive, since with increasing landmark items, it becomes more likely to satisfy fairness constraints, thereby increasing the overall percentage. 

\subsubsection{Qualitative Study}
\noindent
We conduct a qualitative study with 10 graduate students to compare perceived recommendation quality - a score of 1 (least satisfaction) to 5 (highest satisfaction) - under block-level vs. window-level fairness in two online content recommendation applications:
(1) \textbf{Song playlists}, consisting of 15 tracks from three genres (pop, electronic, and country) displayed over three pages (blocks), and 
(2) \textbf{ML content}, comprising 15 YouTube videos across three types (tutorial, project, and explainer) displayed over three pages (blocks). 

For each task, two recommendation orders---one ensuring per-page (block-level) and the other overall (window-level) representation---are presented in randomized order, with participants blinded to the model used. Participants answer three questions for each application:  
(1) What is your average satisfaction with the list generated by Model A?  
(2) What is your average satisfaction with the  list generated by Model B? 
(3) Between Model A and Model B, which list do you prefer overall? 

\noindent Table~\ref{tab:block-vs-window} summarizes the qualitative feedback from participants comparing block-level and window-level fairness models across the two applications. Overall, participants reported noticeably higher satisfaction with the recommendations produced by the block-level model (average satisfaction above 4.2) compared to the window-level model (average satisfaction around 3.0). A strong majority of users (approximately 89\%) preferred the block-level ordering in both the song playlist and ML content scenarios. These results suggest that enforcing fairness at the block level yields recommendations that are perceived as both fairer and more satisfying to users, particularly in stream settings.

\begin{table}[!htbp]
\centering
\setlength{\tabcolsep}{4pt} 
\renewcommand{\arraystretch}{1.1} 
\begin{tabular}{p{1.5cm} >{\centering\arraybackslash}p{1.5cm} >{\centering\arraybackslash}p{1.5cm} >{\centering\arraybackslash}p{1.5cm} >{\centering\arraybackslash}p{1.5cm}}
\toprule
\textbf{Application} & 
\textbf{Avg.\ Sat.\ (Block)} & 
\textbf{Avg.\ Sat.\ (Window)} & 
\textbf{\% Prefer Blocks} & 
\textbf{\% Prefer Windows} \\
\midrule
Song Playlist & 4.22 & 3.00 & 89\% & 11\% \\
ML Content    & 4.33 & 3.00 & 89\% & 11\% \\
\bottomrule
\end{tabular}
\caption{\small Qualitative study results: block vs.\ window-level fairness}
\label{tab:block-vs-window}
 \vspace{-0.1in}
\end{table}
 \vspace{-0.3in}

\vspace{-0.1in}
\subsection{Running Time of Fairness Algorithms}\label{sec:expqp}

We first present the effectiveness of {\tt Monitor-BFair} and {\tt BFair-ReOrder} varying pertinent parameters. In Section~\ref{sec:bcompqp}, we separately compare implemented baselines with our solutions, exhibiting that the baseline is $2-3$ order of magnitude slower than ours. 

\begin{figure*}[!htbp]
\begin{center}
\begin{minipage}{0.3\textwidth}
\includegraphics[width=1\textwidth]{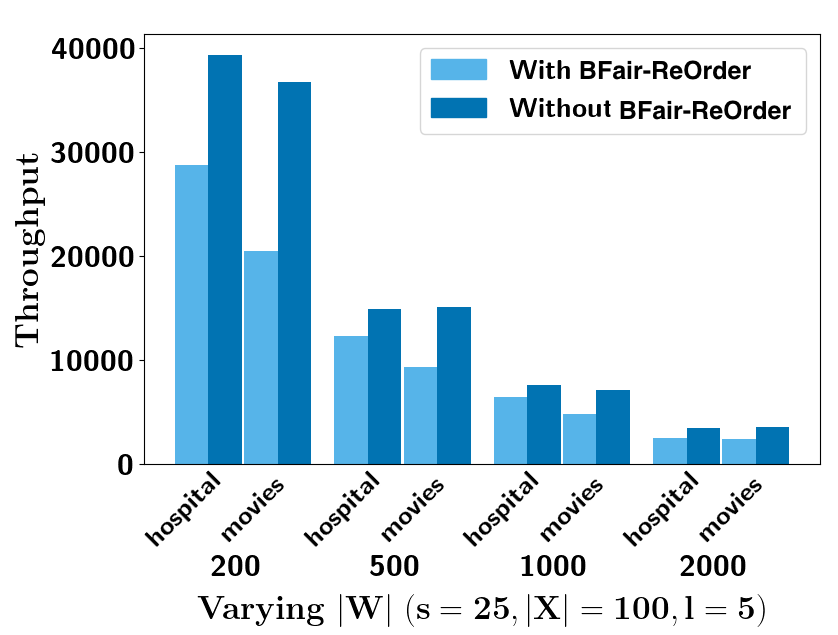}
\subcaption{Varying window sizes}\label{fig:thprtwin}
\end{minipage}
\begin{minipage}{0.3\textwidth}
\includegraphics[width=1\textwidth]{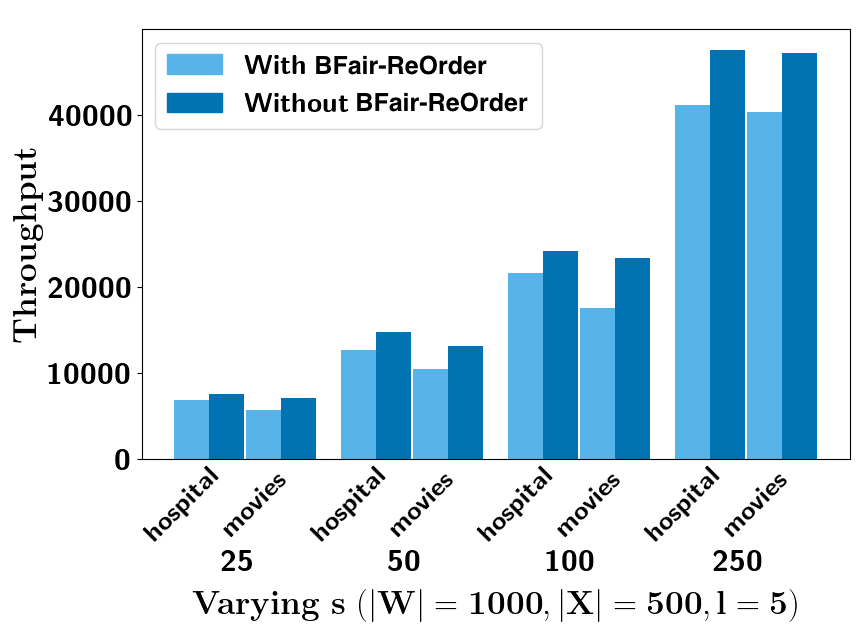}
\subcaption{Varying block sizes}\label{fig:thprtblc}
\end{minipage}
\begin{minipage}{0.29\textwidth}
\includegraphics[width=1\textwidth]{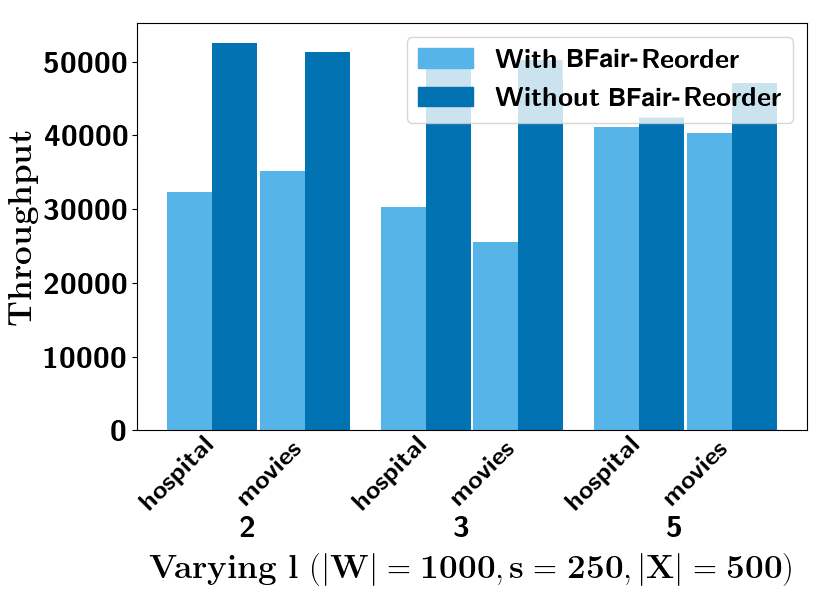}
\subcaption{Varying cardinality}\label{fig:thprtcar}
\end{minipage}
\end{center}
\vspace{-0.2in}
\caption{\small Throughput with and without {\tt BFair-ReOrder}}\label{fig:thprt}
\vspace{-0.1in}
\end{figure*}
\vspace{-0.1in}
\subsubsection{Throughput}
In these experiments, we measure the throughput (per second) of {\tt Monitor-BFair} by varying $|W|$, $s$, and $\ell$, considering both scenarios: one where re-ordering involving landmark items is required using {\tt BFair-ReOrder}, and one where it is not. Figure \ref{fig:thprtwin} presents these results and demonstrates that our algorithms exhibit very high throughout (as high as $40,000$ at cases). As expected, {\tt Monitor-BFair} incurs higher processing time when reordering is required, resulting in lower throughput, and vice versa. We also observe an inverse relationship between $|W|$ and throughput. This is indeed true as the window size increases $|W|$, this leads to  processing higher number of blocks, leading to more time.

\noindent On the other hand, in Figure \ref{fig:thprtblc}, we observe the directly proportional relationship between $s$ and throughput. Growing $s$ decreases the number of blocks for a fixed $|W|$, thus decreasing the time taken for monitoring fairness and increasing the throughput. 

\noindent The impact of $\ell$ is less pronounced (Figure~\ref{fig:thprtcar}). While higher cardinality may increase running time, evenly distributed constraints often lead to more fair blocks, reducing reordering effort. Thus, overall running time stays stable with moderate $\ell$, and throughput remains nearly constant without {\tt BFair-ReOrder}.


\vspace{-0.1in}
\subsubsection{Tail Latency}
\begin{figure*}[!htbp]
\begin{center}
\begin{minipage}{0.25\textwidth}
\includegraphics[width=1\textwidth]{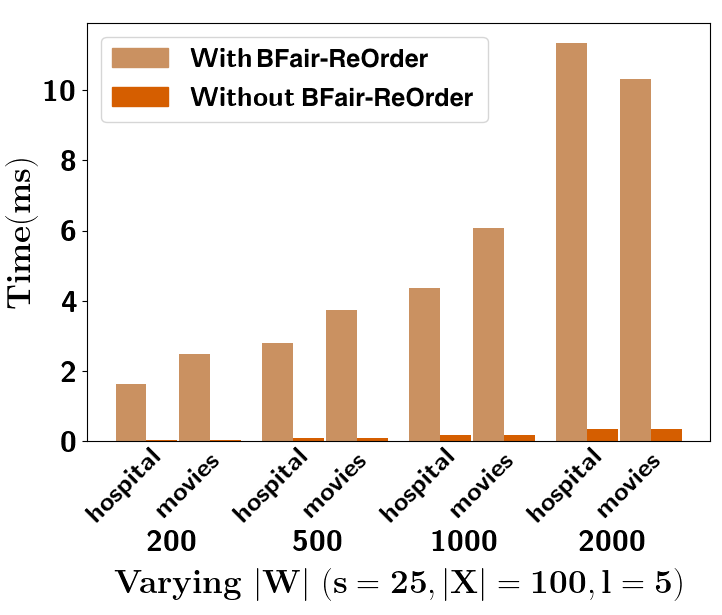}
\subcaption{Varying window sizes}\label{fig:Qproc tail win}
\end{minipage}
\begin{minipage}{0.26\textwidth}
\includegraphics[width=1\textwidth]{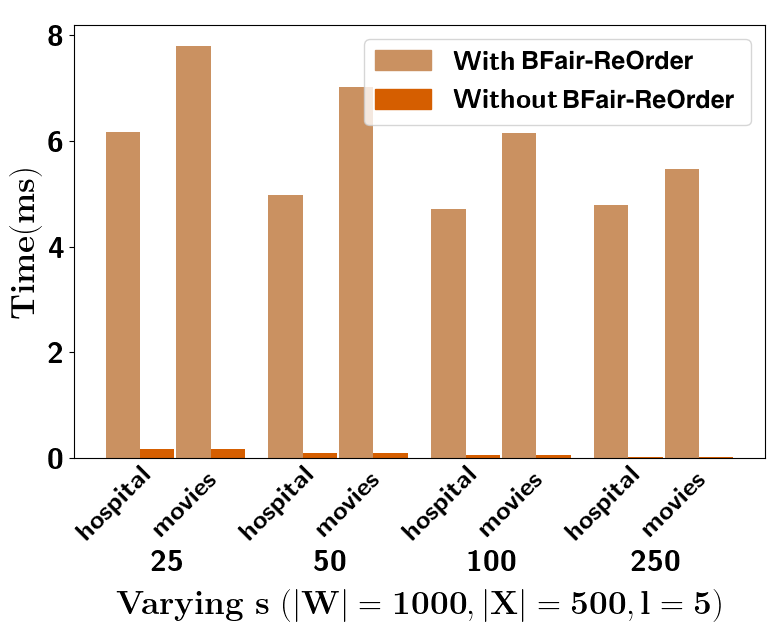}
\subcaption{Varying block Sizes}\label{fig:Qproc tail block}
\end{minipage}
\begin{minipage}{0.25\textwidth}
\includegraphics[width=1\textwidth]{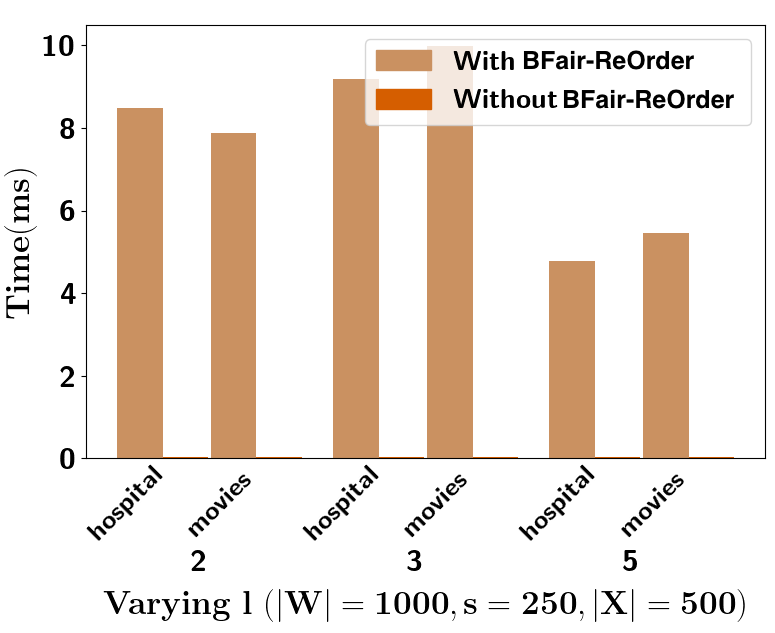}
\subcaption{Varying cardinality}\label{fig:Qproc tail car}
\end{minipage}
\end{center}
\vspace{-0.2in}
\caption{\small 90 percentile running time tail latency over 5000 windows}\label{fig:Qproc tail}
\end{figure*}

We report the 90th percentile tail latency of {\tt Monitor-BFair} over 5,000 windows, both with and without algorithm {\tt BFair-ReOrder}, as shown in Figure~\ref{fig:Qproc tail}.

\noindent As shown in Figure~\ref{fig:Qproc tail win}, tail latency increases with larger $|W|$. For instance, the time to process a single fairness constraint can occasionally exceed 10 ms when $W = 2000$. However, it is important to note that this represents the slowest 10\% of cases and does not reflect the typical running time. Figure~\ref{fig:Qproc tail block} illustrates the expected decrease in tail latency with increasing $s$, consistent with earlier observations. Similarly, Figure~\ref{fig:Qproc tail car} shows the effect of increasing $\ell$. Due to variations in the distribution of values across different $\mathcal{A}$, the tail latency does not exhibit a clear trend in this case.

\noindent Tail latency rises notably when {\tt BFair-ReOrder} is integrated into {\tt Monitor-BFair}, as reordering incurs substantial overhead. Without reordering, latency is minimal.
\subsubsection{Running Time \& Memory Consumption}
Average running time is typically in the order of a fraction of milliseconds, even when reordering is required. Memory consumption of {\tt Monitor-BFair} is negligible and consistent with the sketch memory consumption presented later. Detailed results can be found in our technical report~\cite{git}.

\subsubsection{Running Time compared  to Baseline}\label{sec:bcompqp}  
\begin{figure*}[!htbp]
\begin{center}
\begin{minipage}{0.25\textwidth}
\includegraphics[width=1\textwidth]{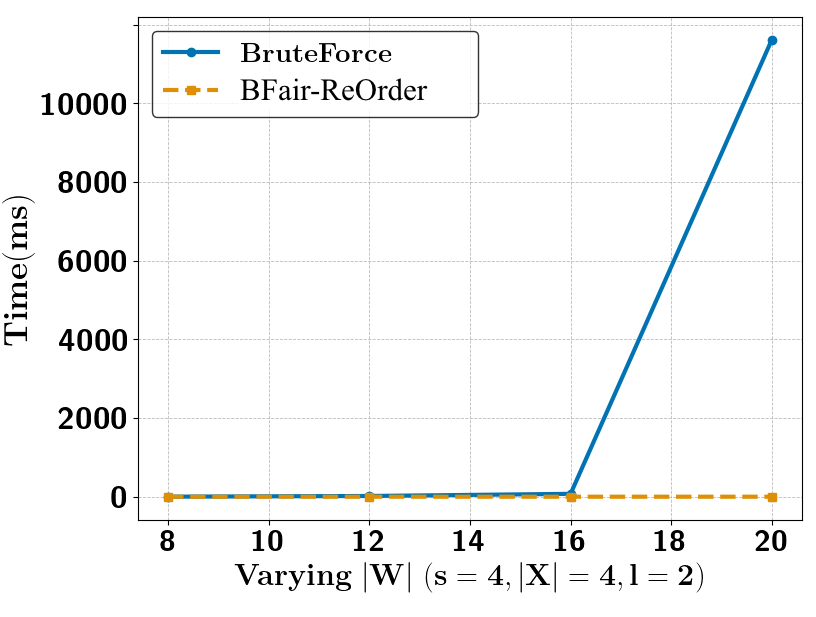}
\subcaption{Varying window sizes}\label{fig:Brute win}
\end{minipage}
\begin{minipage}{0.26\textwidth}
\includegraphics[width=1\textwidth]{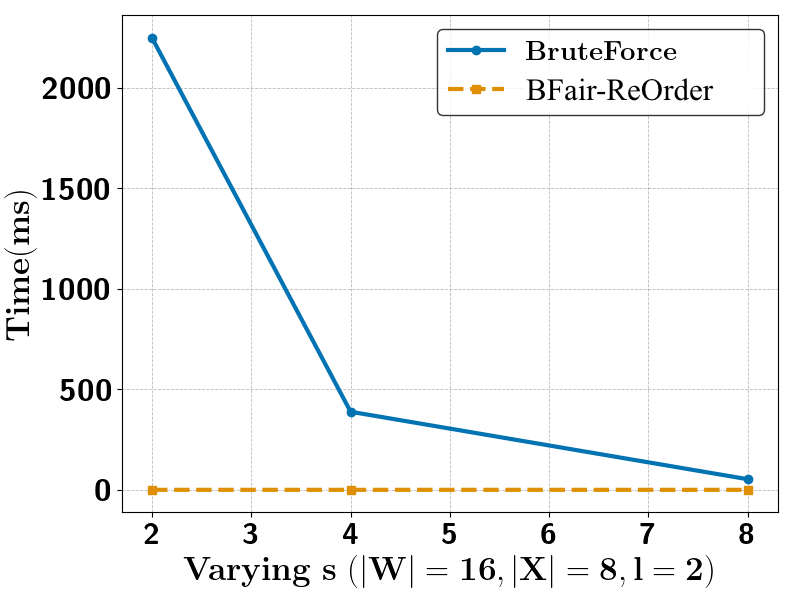}
\subcaption{Varying block sizes}\label{fig:Brute blc}
\end{minipage}
\begin{minipage}{0.25\textwidth}
\includegraphics[width=1\textwidth]{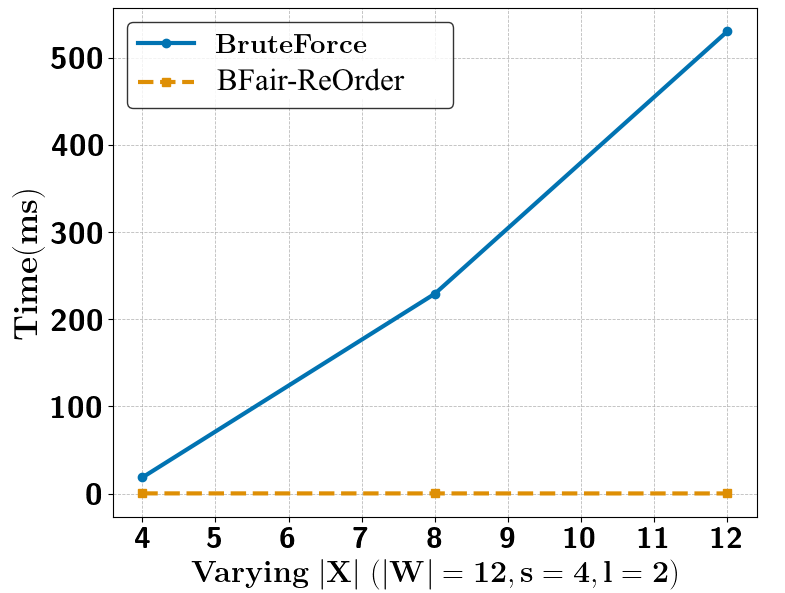}
\subcaption{Varying landmark items}\label{fig:Brute land}
\end{minipage}
\end{center}
\vspace{-0.2in}
\caption{\small {\tt BruteForce} vs {\tt BFair-ReOrder} running time}\label{fig:Brute}
\end{figure*}

We compare the runtime of {\tt BFair-ReOrder} with that of {\tt BruteForce} across varying values of $|W|$, $s$, and $|\mathcal{X}|$ on Hospital dataset, as shown in Figure~\ref{fig:Brute}. Due to the extreme inefficiency of the {\tt BruteForce} algorithm, the streaming environment automatically terminated the process (connection closed due to polling for over 5 minutes) when attempting experiments with window and landmark sizes above 24 items or cardinality beyond 3.

\noindent Figure~\ref{fig:Brute win} shows the inefficiency of {\tt BruteForce}: while comparable to {\tt BFair-ReOrder} for $|W| \leq 16$, its runtime explodes beyond 10 s at $|W| = 20$, confirming its exponential complexity. {\tt BruteForce} exhaustively searches all reorderings, making it highly sensitive to $s$ (Figure~\ref{fig:Brute blc}); larger $s$ reduces blocks, increasing the chance of early fairness. Increasing $|\mathcal{X}|$ similarly extends runtime but with a gentler slope. Overall, {\tt BFair-ReOrder} is over three orders of magnitude faster. {\tt BSketchQP} is consistently outperformed by {\tt Monitor-BFair}, and those results are omitted for brevity.



\begin{figure*}[!htbp]
\begin{center}
\begin{minipage}{0.28\textwidth}
\includegraphics[width=1\textwidth]{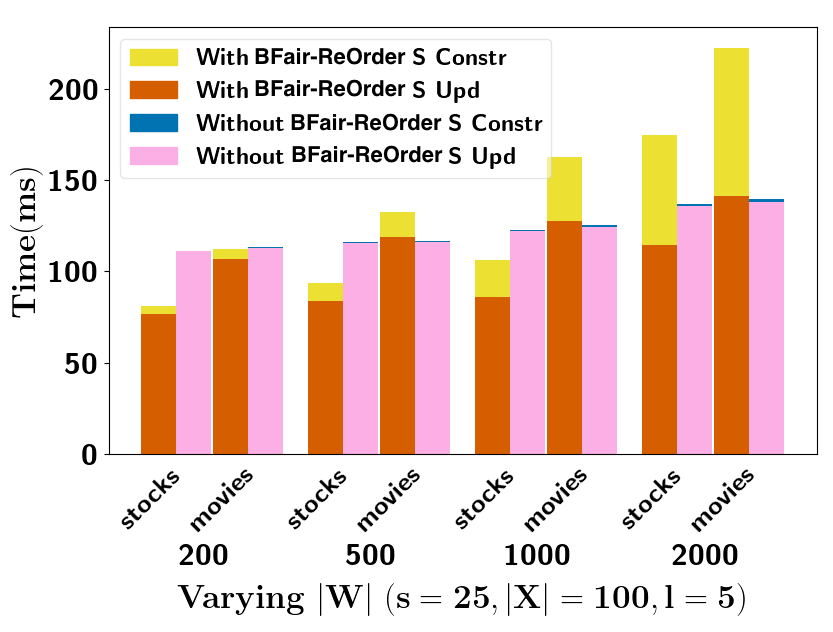}
\subcaption{Varying window sizes}\label{fig:proc win}
\end{minipage}
\begin{minipage}{0.28\textwidth}
\includegraphics[width=1\textwidth]{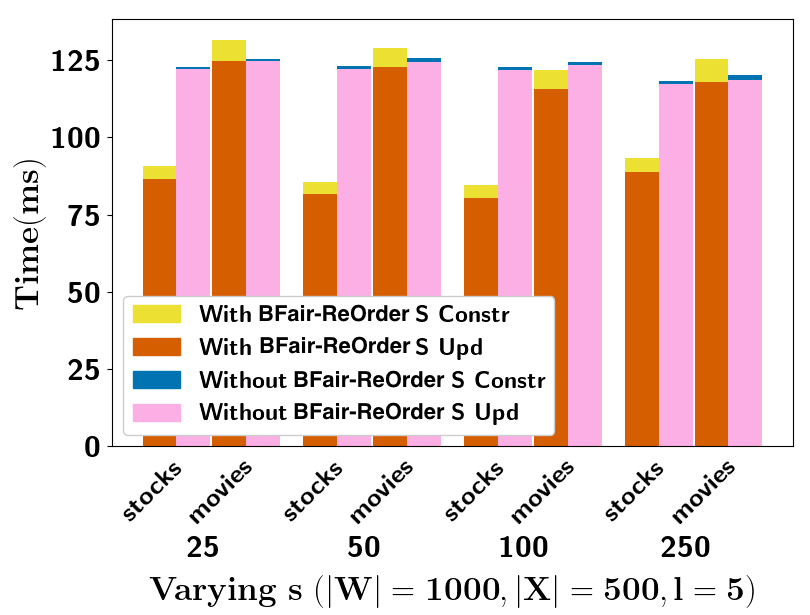}
\subcaption{Varying block sizes}\label{fig:proc blc}
\end{minipage}
\begin{minipage}{0.27\textwidth}
\includegraphics[width=1\textwidth]{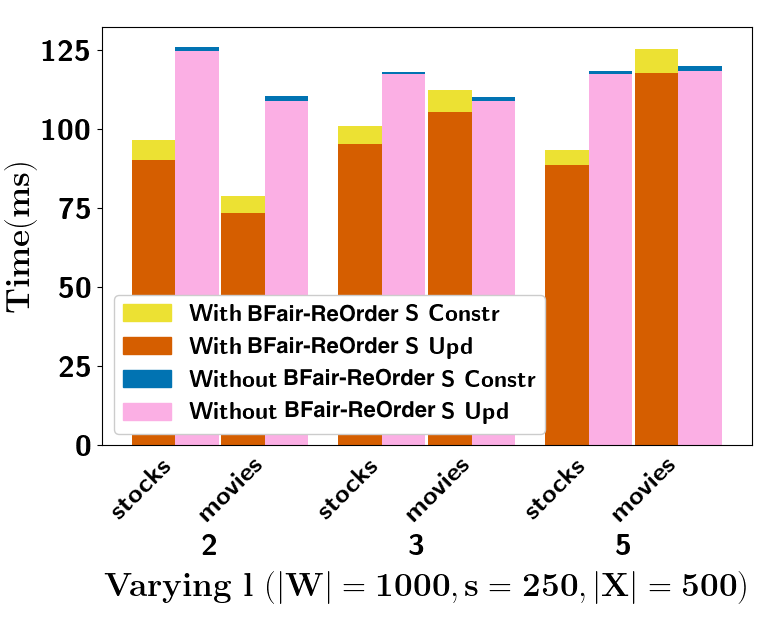}
\subcaption{Varying cardinality}\label{fig:proc car}
\end{minipage}
\end{center}
\vspace{-0.2in}
\caption{\small Total preprocessing running time for 5000 windows}\label{fig:Preprocessing Running Time}
\end{figure*}

\subsection{Preprocessing Analyses}\label{sec:expprep}
We present the preprocessing analyses of our solutions, and later in the section we present the comparison with appropriate baselines.
\vspace{-0.1in}
\subsubsection{Sketch Construction \& Update}
We report the preprocessing time for 5,000 windows on the stocks and movies datasets in Figure~\ref{fig:Preprocessing Running Time}, showing the time breakdown for sketch construction and updates, with and without {\tt BFair-ReOrder} (please note if {\tt BFair-ReOrder} is used the sketch needs to be constructed from scratch). Without {\tt BFair-ReOrder}, sketch construction time is negligible, as it is built once and updated in the remaining windows. 


\noindent As shown in Figure~\ref{fig:proc win}, preprocessing time grows with $|W|$ due to higher sketch construction cost, while update time remains stable. Varying $s$ or $\ell$ has minimal impact (Figure~\ref{fig:proc blc}), showing robustness to block size and cardinality. At times, preprocessing can be slower without {\tt BFair-ReOrder}, since batch memory access via landmark items (with fixed $|X|=500$) reduces overhead compared to loading new records for each window.


 \vspace{-0.1in}
\subsubsection{Memory Requirement}
\begin{figure*}[!htbp]
\begin{center}
\begin{minipage}{0.25\textwidth}
\includegraphics[width=1\textwidth]{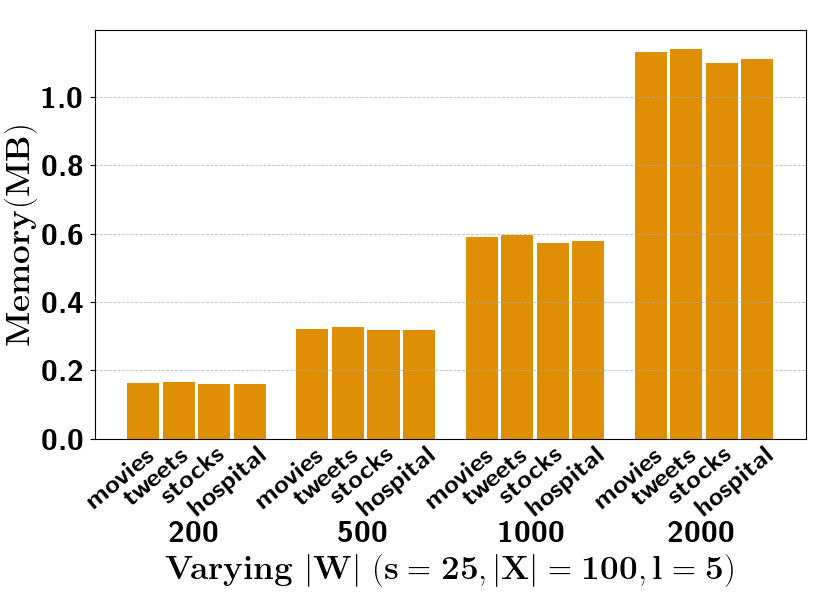}
\subcaption{Varying window size}\label{fig:pre mem win}
\end{minipage}
\begin{minipage}{0.26\textwidth}
\includegraphics[width=1\textwidth]{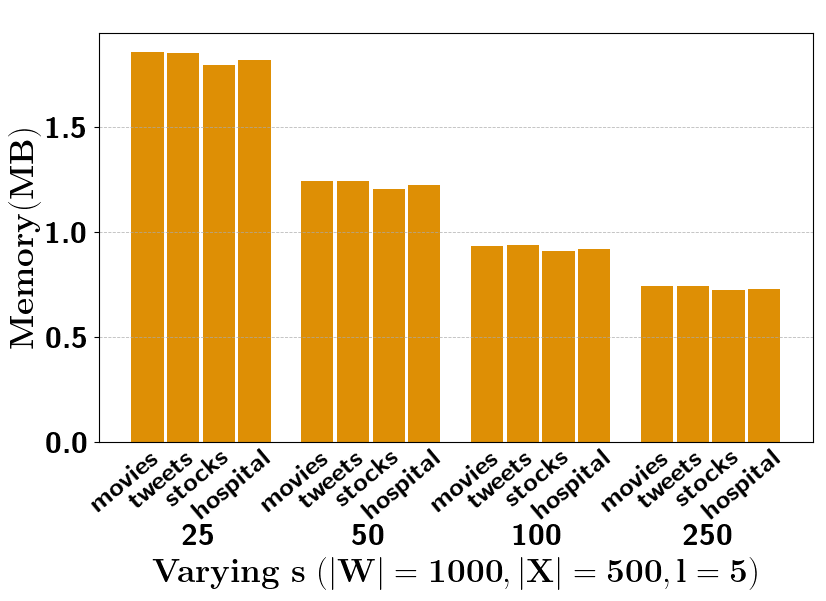}
\subcaption{Varying block size}\label{fig:pre mem blc}
\end{minipage}
\begin{minipage}{0.25\textwidth}
\includegraphics[width=1\textwidth]{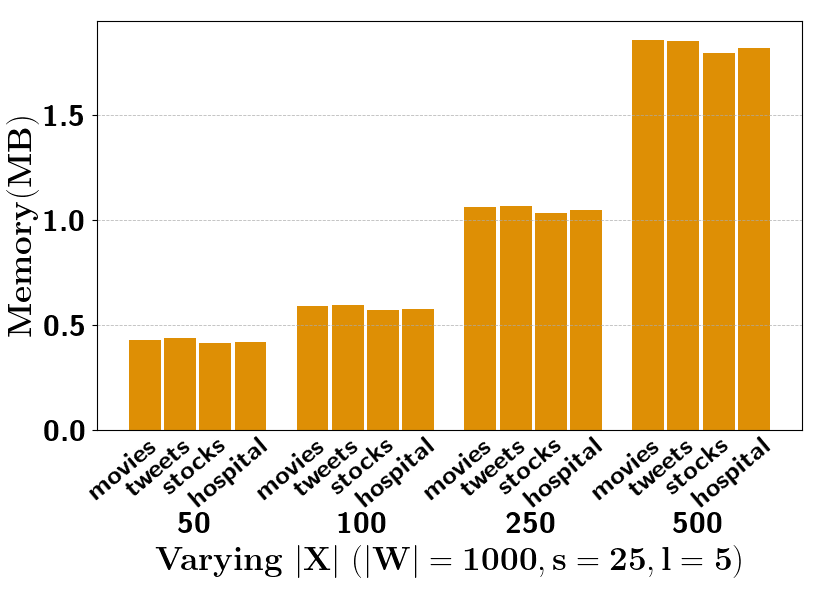}
\subcaption{Varying landmark items}\label{fig:pre mem land}
\end{minipage}
\end{center}
\vspace{-0.2in}
\caption{\small Average peak memory consumption over 5000 windows}\label{fig:Preprocessing Memory Consumption}
\end{figure*}

\noindent Figure~\ref{fig:Preprocessing Memory Consumption} shows average peak memory usage during preprocessing over 5,000 windows, reflecting the maximum memory used—including background processes—under varying parameters.

\noindent As expected, memory consumption of {\tt FSketch} increases with $|W|$ (Figure~\ref{fig:pre mem win}) due to larger sketches. It decreases with increasing $s$ (Figure~\ref{fig:pre mem blc}), as larger blocks reduce the number of configurations needed during {\tt BFair-ReOrder}, lowering memory usage. Similarly, memory usage increases with $|\mathcal{X}|$, not due to sketch size, but because reordering longer streams in {\tt BFair-ReOrder} incurs greater background memory overhead. These experiments corroborate our theoretical analyses, exhibiting that proposed {\tt FSketch} is lean and bounded by window size.
 \vspace{-0.1in}
\subsubsection{Comparison with Baseline}
\begin{figure*}[!htbp]
\begin{center}
\begin{minipage}{0.25\textwidth}
\includegraphics[width=1\textwidth]{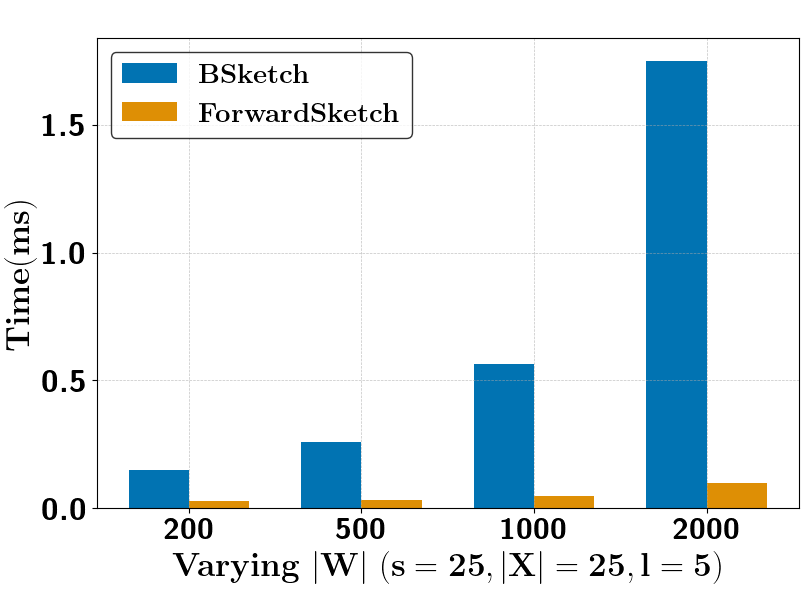}
\subcaption{Varying window size}\label{fig:Bsketch win}
\end{minipage}
\begin{minipage}{0.26\textwidth}
\includegraphics[width=1\textwidth]{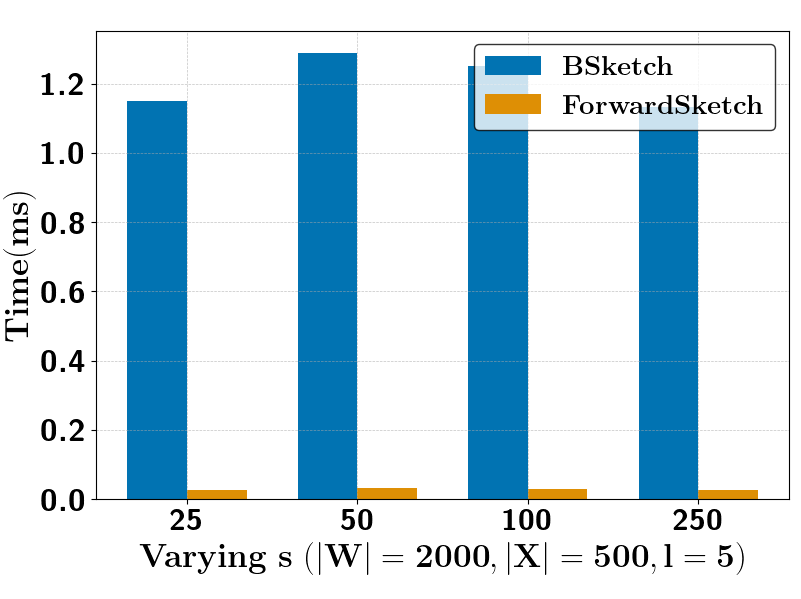}
\subcaption{Varying block size}\label{fig:Bsketch blc}
\end{minipage}
\begin{minipage}{0.25\textwidth}
\includegraphics[width=1\textwidth]{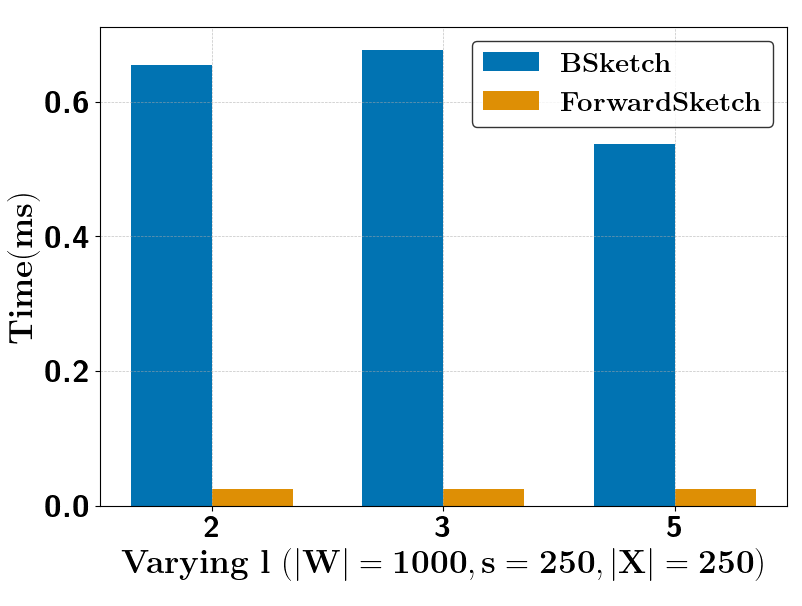}
\subcaption{Varying cardinality}\label{fig:Bsketch car}
\end{minipage}
\end{center}
\vspace{-0.2in}
\caption{\small Preprocessing time of {\tt Backward Sketch} vs {\tt Forward Sketch}}\label{fig:BSketch vs Fsketch}
\end{figure*}

We compare the preprocessing times of \textit{BSketch} and \textit{FSketch} on the stocks dataset (Figure~\ref{fig:BSketch vs Fsketch}). \textit{FSketch} is consistently faster—often by orders of magnitude—with the gap widening as window size grows (Figure~\ref{fig:Bsketch win}), since \textit{BSketch} rebuilds from scratch while \textit{FSketch} incrementally updates. Block size and cardinality have minimal effect (Figures~\ref{fig:Bsketch blc}–\ref{fig:Bsketch car}), with minor variations due to platform level fluctuation.


\begin{figure*}[htbp]
    \centering
    \includegraphics[width=1\linewidth]{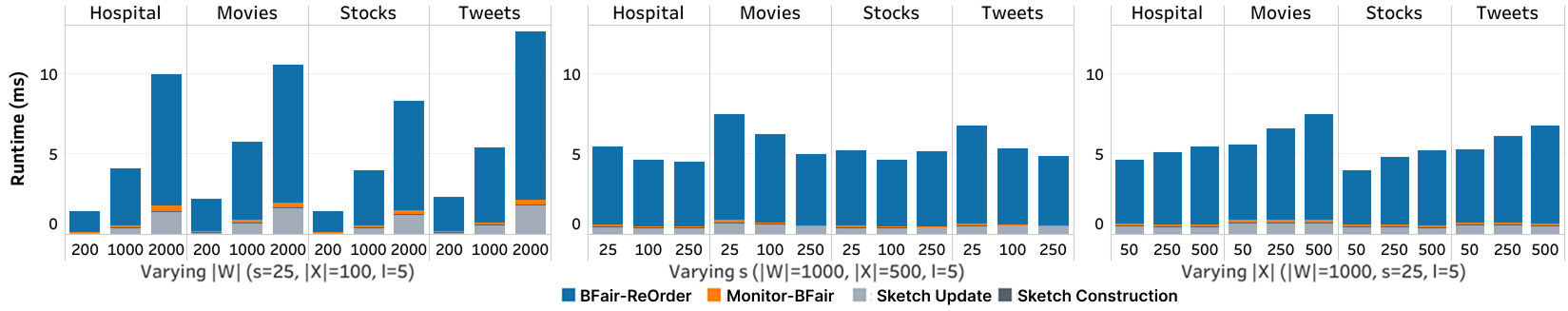} 
    \vspace{-0.25in}
    \caption{\small Component-wise Running Time}
    \label{fig:overhead}
\end{figure*}

\vspace{-0.05in}
\subsection{Component-wise Running Time}\label{sec:overhead}
We evaluate the average runtime of four core components — Sketch Construction, Sketch Update, {\tt BFair-ReOrder}, and {\tt Monitor-BFair} — on all four dataset under varying $|W|$, $s$, and $|\mathcal{X}|$. {\tt BFair-ReOrder} (Figure~\ref{fig:overhead}) incurs the highest time as expected, followed by Sketch Construction, while Sketch Update and {\tt Monitor-BFair} remain consistently lightweight. Sketch Construction scales linearly with $|W|$ but is largely insensitive to $s$ and $|\mathcal{X}|$. In contrast, {\tt BFair-ReOrder} scales linearly with both $|W|$ and $|\mathcal{X}|$, with runtime slightly decreasing as $s$ increases.
\vspace{-0.1in}

\section{Related Work}\label{sec:related}
To the best of our knowledge, no related work studies the problems we propose. \\
\noindent {\bf Group Fairness.}
Group fairness, initially studied in classification, ensures equitable treatment across demographic groups defined by protected attributes (e.g., race, gender, age), commonly measured via \emph{demographic} or \emph{statistical parity}~\cite{demo,stat}. In data management, it has been extensively explored in ranking and recommendation~\cite{kuhlman2020rank,stoyanovich2020responsible,htun2021perception,pitoura2022fairness,wei2022rank}, where fairness notions such as \emph{top-$k$ parity}~\cite{kuhlman2020rank} and \emph{P-fairness}~\cite{wei2022rank} ensure balanced group representation across results or prefixes. Extending these guarantees to \emph{multiple} protected attributes remains computationally challenging~\cite{islam2022satisfying,campbell2024query} due to the exponential growth of intersecting subgroups.

{\em \noindent We generalize the notion of \emph{P-fairness} to every block (instead of every prefix), and initiate its study in the streaming settings. }

\noindent {\bf Data streams.}
Research on data streams has evolved from efficient processing to maintaining data quality under continuous updates. Foundational systems such as CQL~\cite{ArasuBW06}, C-SPARQL~\cite{BarbieriBCVG09}, and window join algorithms~\cite{GolabO03,LimLLWS06,MunagalaSW07} addressed scalable, memory-efficient stream handling. Subsequent work focuses on consistency and constraint management, with denial constraints~\cite{ChuIP13} enabling logical validation and recent frameworks~\cite{Pena0N24,Papastergios25} supporting real-time monitoring and adaptive schema enforcement. Existing works~\cite{SongZWY15,langhi2025evaluating} also propose automatic repair techniques to preserve constraint satisfaction during streaming updates.



\emph{ \noindent In contrast, we focus on enforcing fairness constraints within each block of the window, allowing only data rearrangement without modifying or imputing values. As a result, existing approaches on data quality and repair are not directly applicable to our setting.}

\noindent {\bf Stream fairness.} Research on data stream have recently started paying more attention to fairness. Baumeister et al.~\cite{BaumeisterFSSW25} checks fairness measures such as demographic parity and equalized odds by estimating probabilities from live data streams.  Wang et al.~\cite{WangSYKZ0ZKSWB023} address fairness in evolving data streams with \textit{Fair Sampling over Stream (FS$^2$)}, a method that balances class distribution while considering fairness and concept drift. They also propose \textit{Fairness Bonded Utility (FBU)}, a metric to trade-off between fairness and accuracy.

\emph{In contrast, we propose a framework that ensures group fairness continuously over data streams by monitoring fairness in real time and intelligently reordering window to maximize the achievable level of fairness.}

\vspace{-0.2in}
\section{Conclusion}
We initiate the study of group fairness on data streams, which enforces fairness at a finer granularity through block-level evaluation within sliding windows. This localized approach captures fairness violations more precisely and flexibly than traditional window-level methods. We propose a comprehensive computational framework that includes an efficient sketch-based monitoring structure and an optimal stream reordering algorithm, both designed to support continuous fairness monitoring and reordering in real time. Together, these contributions enable scalable, low-latency, and principled enforcement of group fairness in high-throughput streaming systems, offering strong theoretical guarantees and practical utility for dynamic, fairness-critical applications.

\noindent We are currently investigating how to extend the proposed framework to simultaneously handle multiple protected attributes (e.g., gender, ethnicity, and age). While we recognize the inherent computational intractability of this problem, our preliminary findings indicate that the reordering problem in particular poses substantial challenges, as the current solution does not generalize to satisfying fairness across multiple attributes.



\bibliographystyle{ACM-Reference-Format}
\bibliography{sample-base}

\end{document}